\documentclass[runningheads]{llncs}
\usepackage[T1]{fontenc}
\usepackage{graphicx}
\usepackage{amssymb}

\usepackage{times}
\usepackage{soul}
\usepackage{url}
\usepackage[utf8]{inputenc}
\usepackage{caption}
\usepackage{amsmath}
\usepackage{booktabs}
\usepackage[switch]{lineno}

\newcommand{\sbt}{\,\begin{picture}(-1,1)(-1,-3)\circle*{2}\end{picture}\ }

\usepackage{xcolor}

\usepackage{booktabs}

\usepackage{multirow}
\usepackage{wrapfig}

\usepackage{setspace}

\usepackage{float}
\floatstyle{plaintop}
\restylefloat{table}

\usepackage{tablefootnote}

\usepackage{subcaption}
\usepackage{lipsum}


\usepackage{algorithm,algpseudocode}
\algnewcommand{\REQUIRE}[1]{%
  \State \textbf{Require:}
  \Statex \hspace*{\algorithmicindent}\parbox[t]{.8\linewidth}{\raggedright #1}
}

\algnewcommand{\Inputs}[1]{%
  \State \textbf{Inputs:}
  \Statex \hspace*{\algorithmicindent}\parbox[t]{.8\linewidth}{\raggedright #1}
}
\algnewcommand{\Outputs}[1]{%
  \State \textbf{Output:}
  \Statex \hspace*{\algorithmicindent}\parbox[t]{.8\linewidth}{\raggedright #1}
}
\algnewcommand{\Initialize}[1]{%
  \State \textbf{Initialize:}
  \Statex \hspace*{\algorithmicindent}\parbox[t]{.8\linewidth}{\raggedright #1}
}

\algnewcommand{\algorithmicforeach}{\textbf{for each}}
\algdef{SE}[FOR]{ForEach}{EndForEach}[1]
  {\algorithmicforeach\ #1\ \algorithmicdo}
  {\algorithmicend\ \algorithmicforeach}

\algrenewcommand\textproc{}

\usepackage{array}
\usepackage{makecell}

\usepackage{calrsfs}
\DeclareMathAlphabet{\pazocal}{OMS}{zplm}{m}{n}
\SetMathAlphabet\pazocal{bold}{OMS}{zplm}{bx}{n}

\usepackage{hyperref}
\hypersetup{
    colorlinks=true,
    linkcolor=blue,
    filecolor=magenta,
    citecolor=violet,
    urlcolor=cyan,
    pdftitle={Overleaf Example},
    pdfpagemode=FullScreen,
    }

\begin{document}
%
\title{A Non-Monolithic Policy Approach of Offline-to-Online
Reinforcement Learning}
%
%

\author{JaeYoon Kim\inst{1}\orcidID{0000-0003-0898-5097} \and
Junyu~Xuan\inst{1}\orcidID{0000-0002-8367-6908} \and
Christy~Liang\inst{2}\orcidID{0000-0001-7179-5208}  \and
Farookh~Hussain\inst{1}\orcidID{0000-0003-1513-8072}}

\authorrunning{JaeYoon Kim et al.}
%
\institute{Australian Artificial Intelligence Institute (AAII), University of Technology Sydney, Australia
\email{JaeYoon.Kim@student.uts.edu.au, \{Junyu.Xuan,\;Farookh.Hussain\}@uts.edu.au}\\
\and
Visualisation Institute, University of Technology Sydney, Australia\\
\email{Jie.Liang@uts.edu.au}}
\maketitle              
\begin{abstract}
Offline-to-online reinforcement learning (RL) leverages both pre-trained offline policies and online policies trained for downstream tasks, aiming to improve data efficiency and accelerate performance enhancement. An existing approach, Policy Expansion (PEX), utilizes a policy set composed of both policies without modifying the offline policy for exploration and learning. However, this approach fails to ensure sufficient learning of the online policy due to an excessive focus on exploration with both policies. Since the pre-trained offline policy can assist the online policy in exploiting a downstream task based on its prior experience, it should be executed effectively and tailored to the specific requirements of the downstream task. In contrast, the online policy, with its immature behavioral strategy, has the potential for exploration during the training phase. Therefore, our research focuses on harmonizing the advantages of the offline policy, termed exploitation, with those of the online policy, referred to as exploration, without modifying the offline policy. In this study, we propose an innovative offline-to-online RL method that employs a non-monolithic exploration approach. Our methodology demonstrates superior performance compared to PEX. The code for this comparison is readily available. \href{https://github.com/jangikim2/Offline-to-online-RL-with-non-monolithic-exploration-methodology/tree/main}{Source-Code}
\keywords{Non-Monolithic Exploration \and Offline-to-Online Reinforcement
Learning \and Mode-Switching Controller.}
\end{abstract}

\section{Introduction}
Offline RL \cite{129} has been researched to address the disadvantages of online RL, such as the non-trivial cost associated with collecting data from a real environment and the risks of physical interactions with the environment, such as with real robots \cite{134,135}. However, it still suffers from limited performance due to sub-optimal datasets in downstream tasks.

Offline-to-online RL addresses the disadvantages of both offline RL and online RL \cite{131,136}. It involves pre-training a policy on an offline dataset, followed by fine-tuning the online policy using the pre-trained offline policy in a real environment.

However, two main issues arise in offline-to-online RL \cite{137}. The first issue stems from offline RL and involves out-of-distribution (OOD) challenges due to distribution mismatch, which impacts the exploitation phase during the fine-tuning process. The second issue pertains to the exploration aspect of offline-to-online RL being hindered by the constrained policy of offline RL. 

To address these persistent issues, numerous studies have been conducted within the RL community \cite{130,133,139}. Among these approaches, Policy Expansion (PEX) \cite{132}, the current state-of-the-art offline-to-online RL algorithm that utilizes an unmodified offline policy, leverages a policy set comprising both policies without altering the offline policy in the pre-training-to-fine-tuning scheme for exploration and learning. However, the excessive involvement of both policies in exploration leads to deficiencies in both exploitation and exploration. Its step operation unit cannot ensure the autonomous training of the online agent, ultimately deteriorating the performance of PEX.

In a pre-training-to-fine-tuning scheme, the offline policy is pre-trained using an offline dataset, proving beneficial for exploiting the distribution of a downstream task. The unmodified offline policy in offline-to-online RL can also assist its counterpart, the online policy, in guiding exploration to some extent. Thus, the unmodified offline policy becomes crucial for effectively exploiting the downstream task. Simultaneously, the online policy has the potential to explore a downstream task due to its immature policy, addressing the distribution shift between the offline dataset and the downstream task. Achieving optimal agent performance requires ensuring proper execution timing and duration for the online policy in relation to the offline policy. Therefore, our research focuses on how to reconcile the advantages of the offline policy (exploitation) and the online policy (exploration) in offline-to-online RL, without compromising the integrity of the offline policy, to enhance overall agent performance.

In this research, we propose a novel model for offline-to-online RL that utilizes a non-monolithic exploration methodology without modifying the offline policy. In our model, an offline policy and an online policy are specialized in exploitation and exploration, respectively. Thus, our offline-to-online model features a policy set similar to PEX, complemented by a mode-switching controller to select an active policy. An agent utilizing this model without modifying the original offline policy should address the following questions: How does the agent leverage the offline policy for exploitation based on the nature of the downstream task? Additionally, how does the agent effectively explore with the online policy?

The main contributions of our research are outlined below:
\begin{itemize} 
\item {The development of our offline-to-online RL model leverages a non-monolithic exploration methodology to strategically focus on both exploitation and exploration.}
\item  {The ability to leverage flexibility and generalization in a downstream task.}
\end{itemize}

\section{Related work} \label{Related work}

\subsection{Non-monolithic exploration}
A noise-based monolithic exploration RL agent capitalizes on a final action by adding its policy's action to random noise for exploration. In contrast, the non-monolithic exploration methodology comprises discrete agents—one for exploitation and another for exploration—managed by a mode-switching controller that selects the appropriate agent at the right time. Fig. 1, shown in `supplementary document.pdf' on the GitHub mentioned in the Abstract, illustrates the diagrams for both the noise-based monolithic exploration RL agent and the non-monolithic exploration RL agent.

The significance of `when' to explore, a primary consideration in the non-monolithic exploration methodology, has been emphasized in \cite{127,128,63}, contrasting with the traditional focus on `how' to explore, typically associated with monolithic exploration approaches. In \cite{63}, the non-monolithic exploration methodology incorporates `Homeostasis' \cite{64} as a mode-switching controller, abbreviated as `Homeo' (see `A.5 Homeostasis' in `supplementary document.pdf'), to select an active agent between exploitation and exploration. `Homeo' leverages the `value promise discrepancy', $D_{promise}(t-k,t)$, which represents the difference in the value function over k steps:

\begin{equation}
\label{eq:valuepromise}
D_{promise}(t-k,t) := \; \big | V(s_{t - k}) - \sum_{i=0}^{k-1}\gamma^{i}R_{t-i}  -  \gamma^{k}V(s_{t}) \big |
\end{equation}
where $V(s)$ is an agent's value estimate at state s, $R$ is a reward and $\gamma$ is a discount factor.

\subsection{Pre-training}
Pre-training is designed to enhance sample efficiency and transfer prior knowledge to a downstream task \cite{156}. The main types include online or offline pre-training, supervised or unsupervised pre-training \cite{84}, and representation \cite{162} or policy pre-training \cite{163}.

Our research focuses on offline pre-training in RL, which involves utilizing an off-policy approach, as seen in offline RL methods such as CQL \cite{141}, IQL \cite{142}, AWAC \cite{157}, and COMBO \cite{158}.

\subsection{Offline RL} 
Online RL presents significant challenges due to the high costs and risks associated with interactions in real environments. Consequently, the RL research community has shown interest in Offline RL, which leverages existing pre-used datasets. However, a persistent issue is the vulnerability to distributional shifts between the offline dataset and the distribution of the downstream task. Offline RL has been explored in various areas, including policy constraints, importance sampling, regularization, uncertainty estimation, model-based methods, one-step methods, imitation learning, and trajectory optimization \cite{146,145}.

In the evaluation and improvement of an offline policy, the policy loss term typically includes loss terms related to policy constraints \cite{147,148}, uncertainty \cite{154,155}, and regularization \cite{141,151}.

\subsection{Offline-to-online RL} 
Prior work in offline-to-online RL has focused on consolidating an acquired online policy based on an offline policy \cite{133,137,159}. 

PEX was designed to deviate from the traditional unified framework, thereby avoiding compromises to a pre-trained offline policy. In PEX, $\pi_{\beta}$ is kept frozen, while $\Pi$ consists of both $\pi_{\beta}$ and a learnable policy $\pi_{\theta}$.

\begin{equation}
\label{eqn:PEX_equation_1}
\Pi = [\pi_{\beta}, \pi_{\theta}]
\end{equation}A proposal actions set, $\mathbb{A}\,=\, \left\{ a_{i}\,\sim\,\pi_{i}(s)  | \pi_{i} \in \Pi \right\} $, is formed from $\Pi$ at the current state $s$. One of them will be selected from this set using a categorical distribution, $P_{\textbf{w}}$, which is associated with a value function $\textbf{Q}_{\phi}\,=\, \left[ Q_{\phi}(s,a_{i})  | a_{i} \in \mathbb{A} \right] \in \mathbb{R}^{K}$ where $K$ is cardinality of $\Pi$ (here $K = 2$).

\begin{equation}
\label{eqn:PEX_equation_2}
P_{\textbf{w}}[i] = \dfrac{\text{exp}(Q_{\phi}(s,a_{i})/\alpha)}{\sum_{j}\text{exp}(Q_{\phi}(s,a_{j})/\alpha)}, \hspace{0.2cm}    \forall i \in [1, 	\sbt \,\, \sbt \,\, \sbt \,\, K]
\end{equation} where $\alpha$ is temparature. Then \textbf{w} is selected through $\textbf{w} \sim  P_{\textbf{w}}$. Finally, the composite policy $\tilde{\pi}$ is used for a final action.

\begin{equation}
\label{eqn:PEX_equation_3}
\tilde{\pi}(a|s) = [\delta_{a \sim \pi_{\beta}(s)},\delta_{a \sim \pi_{\theta}(s)}]\textbf{w},\;\;\;\; \textbf{w} \sim  P_{\textbf{w}}
\end{equation} where $\textbf{w} \in \mathbb{R}^{K}$ is a one-hot vector and $\delta_{a \sim \pi}$ is the Dirac delta distribution centered at an action $a \sim \pi$.

Our research does not utilize the categorical distribution $P_{\textbf{w}}$ for exploration and learning.

\section{Our methodology} \label{Our methodology}
The ultimate goal of offline-to-online RL is to leverage the pre-learned knowledge of the offline policy and the adaptability of the online policy. Unlike the unified framework, which compromises the integrity of a pre-trained offline policy, it becomes challenging to timely extract the original knowledge from the offline policy. Conversely, an offline-to-online RL scheme that retains an unmodified offline policy can achieve its objectives if equipped with a suitable mode-switching controller. Although PEX employs such a scheme, it struggles to ensure sufficient online training for exploration and learning, as indicated in Eq. (\ref{eqn:PEX_equation_2}) and Eq. (\ref{eqn:PEX_equation_3}).

\subsection{\textcolor{black}{Offline-to-online RL with a mode-switching controller}}
\begin{figure*}
  \vspace*{-5mm} 
  \centering
  \includegraphics[scale=0.4]{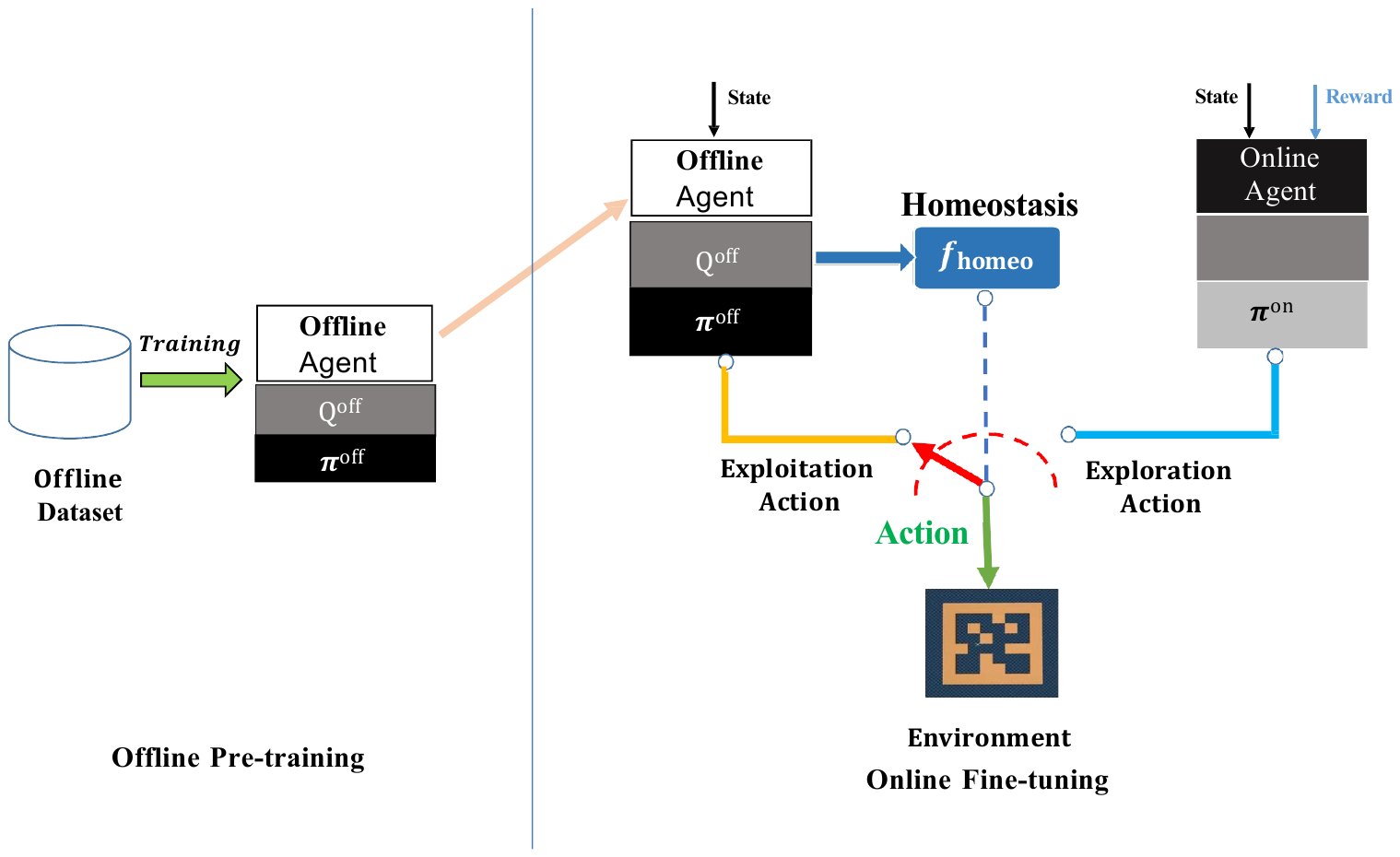}
  \vspace*{-15mm}
  \caption{Illustration of offline-to-online RL training schemes employed in our model with an unmodified offline policy}
  \label{fig:both_architecture}
\end{figure*}

Our model incorporates a non-monolithic exploration methodology into offline-to-online RL, featuring a policy set $\Pi$ comprised of $\pi^{\text{off}}\,\text{and}\,\pi^{\text{on}}$, as illustrated in Fig. \ref{fig:both_architecture}. 

\begin{equation}
\label{eqn:PEX_equation_4}
\Pi = [\pi^{\text{off}}, \pi^{\text{on}}]
\end{equation}

The composite policy $\tilde{\pi}$, based on $\Pi$, consists of both $\pi^{\text{off}}$ and $\pi^{\text{on}}$, with their roles separated into exploitation and exploration, respectively. The activation of $\pi^{\text{off}}$ and $\pi^{\text{on}}$ is determined by a mode-switching controller. We use Homeo, whose algorithm is defined in \cite{63}, as the mode-switching controller in our model, as detailed in `Algorithm 1' of `supplementary document.pdf'. The $D_{\text{promise}}$ function\textsuperscript{\ref{eq:valuepromise}} of the Homeo, denoted as $f_{\text{homeo}}$, utilizes $Q^{\text{off}}(s, Action)$, the state-action value function of $\pi^{\text{off}}$, to monitor changes in the pre-trained $Q^{\text{off}}$ during a predefined timeframe. The decision of $Action$ of $\tilde{\pi}$ as determined by Homeo is as follows:
\begin{align}
\begin{split}
D_{\text{promise}}^{\text{off}} & = D_{\text{promise}}^{\text{off}}(t-k,t)\\
\tilde{\pi}(Action|s) & = \begin{cases}
    a_t^{\text{off}} \sim \pi^{\text{off}}, & \text{for } f_{\text{homeo}}(D_{\text{promise}}^{\text{off}}) = 0\\
    a_t^{on} \sim \pi^{on}, & \text{for } f_{\text{homeo}}(D_{\text{promise}}^{\text{off}}) \not = 0\\
    \end{cases}
\end{split}
\end{align}
where $f_{\text{homeo}}(D_{\text{promise}})$ with the input of $D_{\text{promise}}^{\text{off}}(t-k,t)$ using $Q^{\text{off}}(s,a)$ denotes Homeo for the decision of mode switching controller in which $D_{\text{promise}}^{\text{off}}(t-k,t)$ denotes the value promise discrepancy of $\pi^{\text{off}}$ and the result value of $f_{\text{homeo}}(D_{\text{promise}})$ is based on Bernoulli distribution.

The policy $\tilde{\pi}$ operates with a union replay buffer $\mathrm{D}_{\text{off}} \cup \mathrm{D}_{\text{on}}$. Consequently, $\pi^{\text{on}}$ also utilizes the same union replay buffer for its own training, whether $\pi^{\text{off}}$  or $\pi^{\text{on}}$ is active. This union replay buffer facilitates the rapid improvement of $\pi^{\text{on}}$'s performance.
The framework with discrete agents, $\pi^{\text{off}}$ and $\pi^{\text{on}}$, ensures the primary objective of our research, where $\pi^{\text{off}}$ focuses on exploitation and  $\pi^{\text{on}}$ on exploration. However, the role of $\pi^{\text{on}}$ will change during the late online fine-training stage, a topic that will be discussed later.

\begin{algorithm}[!t]
  \caption{The main algorithm of online stage}
  \label{alg:the_alg}
  \begin{algorithmic}[1]
    \Initialize{\strut {Set the value of \textit{explore\_fixed\_steps}}\\
    {Set the value of \textit{update\_timestep}}\\    
    {Set the value of $\rho$ in (0.0001 $\sim$ 0.9) for Homeo}\\
    {Set $\pi^{\text{off}}$ and $Q^{\text{off}}$ through a pre-trained offline policy and then freeze them}\\}
    
    \Procedure{Evaluate $\pi^{\text{off}}$}{$D_{\text{promise}}\;\text{of}\;\pi^{\text{off}}$}
        \State {Compute\;the\;output\;of\;Homeo\;with\;$D_{\text{promise}}$\;of\\\;\;\;\;\;\;\;\;$\pi^{\text{off}}$\;based\;on\;\textit{explore\_fixed\_steps}}
    \EndProcedure   
    
    \For {$t=0,\ldots,T-1$}
        \If{\textit{update\_timestep}}
                \State {Compute\; $D_{\text{promise}}$ \;of\;$\pi^{\text{off}}$\;through\;$Q^{\text{off}}$}
                \State {Evaluate $\pi^{\text{off}}(D_{\text{promise}}\;\text{of}\;\pi^{\text{off}})$}
            \EndIf
            \If{\textit{output of Homeo == exploitation mode} }
                \State {\textit{Action} $\gets \pi^{\text{off}}$}
            \Else
            \State {\textit{Action} $\gets \pi^{\text{on}}$}
            \EndIf
        \State {Execute \textit{Action}, observe reward and\,next\_state}        
    \EndFor        
  \end{algorithmic}
\end{algorithm}	

Our model adopts a heterogeneous temporal structure for mode-switching exploration. Two agents, each with a distinct purpose, monitor $Q^{\text{off}}$ over a specific period. If the state of $Q^{\text{off}}$ deteriorates within a predefined timeframe, the mode of policy $\tilde{\pi}$ transitions from exploitation to exploration. What triggers this switch? The mode-switching controller recognizes when the current exploitation by the offline policy is no longer reliable and consequently terminates it. This initiates an exploration phase for the downstream task, leading to a shift in $\tilde{\pi}$'s mode to exploration. Control of action is then transferred from $\pi^{\text{off}}$ to $\pi^{\text{on}}$. 

Meanwhile, during the online fine-training, we hypothesize that the role of $\pi^{\text{on}}$ on $\Pi$ changes as follows. During the beginning and middle of online fine-training,
\begin{equation}
\begin{gathered}
\label{entropy_equation_1}
\textsc{$\textit{H}\Bigl(\pi^{off}$}\Bigr) <
\textsc{$\textit{H}\Bigl(\pi^{on}$}\Bigr),\\
\text{therefore},\,\pi^{\text{on}}\,:\,\text{exploration\,policy}
\end{gathered}
\end{equation} where $\textsc{$\textit{H}\Bigl(\cdot$}\Bigr)$ denotes the overall entropy of a policy.
In the late online fine-training stage,
\begin{equation}
\begin{gathered}
\label{entropy_equation_2}
\mathrm{D}_{\text{off}}\,\cup\, \mathrm{D}_{\text{on}}\,\approx\,\mathrm{D}_{\text{Down,OPT}},\\
\text{so},\textsc{$\textit{H}\Bigl(\pi^{off}$}\Bigr) \ge
\textsc{$\textit{H}\Bigl(\pi^{on}$}\Bigr),\\
\text{therefore},\,\pi^{\text{on}}\,:\,\text{exploitation-oriented\,policy}.
\end{gathered}
\end{equation} During the beginning and middle of online fine-training, $\pi^{\text{on}}$ primarily focuses on exploration. However, as training progresses, $\pi^{\text{on}}$ becomes increasingly specialized in the downstream task. Consequently, in the late stages of online training, the Homeo controller selects between the exploitation policy $\pi^{\text{off}}$ and the now exploitation-oriented policy $\pi^{on}$ to enhance exploitation.

PEX also employs a heterogeneous temporal structure in mode-switching exploration but utilizes a different switching mode controller. This controller assesses the value function $Q_{\phi}$ of both agents not just during a specific period but at every training step, as illustrated in Equations (\ref{eqn:PEX_equation_2}) and (\ref{eqn:PEX_equation_3}). Consequently, PEX considers both agents for the combined roles of exploitation and exploration, making these its primary focus at each training step.  To align with PEX's main objectives, $\pi_{\theta}$ should undergo sufficient training during a designated period, with an emphasis on exploring through $\pi_{\theta}$.

In PEX, frequent interventions of $\pi_{\beta}$ within $\tilde{\pi}$ can negatively affect the exploration capabilities of $\pi_{\theta}$. Our model addresses this concern by focusing exclusively on $Q^{\text{off}}$ during specific periods, which allows $\pi^{\text{on}}$  to improve independently. In this model, $\pi^{\text{off}}$ is solely responsible for exploitation, while $\pi^{\text{on}}$ is dedicated to exploration until the midpoint of the online fine-training. During the late stages of online fine-training, $\pi^{\text{on}}$, now more exploitation-oriented,  shifts its focus towards enhancing exploitation. This non-monolithic exploration approach efficiently meets these specific requirements.

\subsection{The ability to leverage flexibility and generalization on a downstream task}

Equations (\ref{eqn:PEX_equation_2}) and (\ref{eqn:PEX_equation_3}) illustrate the operation of $\tilde{\pi}$ per training step during the online fine-tuning phase of PEX. The \textcolor{black}{premise} for ensuring PEX performance is the sufficient training of the online policy $\pi_{\theta}$, emphasizing exploration. However, the operational mechanism of PEX does not guarantee this due to the step operation. Conversely, in our model, if $\pi^{\text{off}}$ does not perform well during a predefined period (referred to as \textit{update\_timestep}), $\pi^{\text{on}}$  can undergo sufficient sequential training over a designated period (referred to as \textit{explore\_fixed\_steps}) based on its exploration capabilities. This setup ensures the autonomous training of $\pi^{\text{on}}$  during the specified timeframe.

Therefore, our model can modulate the execution times of both $\pi^{\text{off}}$ and $\pi^{\text{on}}$ based on the dissimilarity between the distribution of the offline dataset and that of the downstream task. This allows our model to support a flexible and generalized approach, adapting effectively to various situations in both the offline dataset and the downstream task.


Given that $\pi^{\text{off}}$ is extractable and $\pi^{\text{on}}$ is trainable, controlling the performance of $\tilde{\pi}$ flexibly involves addressing two main concerns. Consequently, the role of the modulator, which comprises \textit{update\_timestep} and \textit{explore\_fixed\_steps}, is crucial for the performance of $\tilde{\pi}$. The characteristics of the modulator enable our model to exhibit either an exploitative or exploratory bias, adapting dynamically to the needs of the task.

An optimal policy can be achieved through the careful negotiation of the modulator, striking a balance between exploitative and exploratory biases. Once the aforementioned parameters are well-tuned, this balance becomes achievable for any downstream task. 

The modulating and mode-switching characteristics of our model provide a robust adaptive capacity for various downstream tasks, a feature absent in PEX. Consequently, our model excels in maximizing both pre-experienced and newly acquired knowledge, effectively adapting to different downstream tasks.

The  table \ref{Comparison_ourmodel_pex} shows the comparison between our model and PEX.

\begin{table}[!t]
   \caption{\textcolor{black}{The comparison between our model and PEX.}}
   \begin{minipage}{\columnwidth}
     \begin{center}
       \textcolor{black}{
       \begin{tabular}{p{6cm}p{2cm}p{1cm}}
         \toprule
         \textbf{Item} & \textbf{Our model} & \textbf{PEX} \\
         \hline
         Operation unit of policy & Period & Step\\
         Switching condition on $\Pi$  & Only $Q^{\text{off}}$ & $Q_{\phi}$\\ 
         Switching-mode controller  & Homeostasis & $P_{\textbf{w}}$\\
         Are the roles of agents on $\Pi$ separated? & Yes & No\\          
         The role of online agent changes?  & Yes & No\\         
         Is the online agent trained autonomously?& Yes & No\\
         A modulator exists? & Yes & No\\
         \bottomrule
       \end{tabular}
       }
     \end{center}
   \end{minipage}
   \label{Comparison_ourmodel_pex}
\end{table}

\section{Experiments} \label{Experiments}
The open-source code of PEX\footnote{\url{https://github.com/Haichao-Zhang/PEX}\label{pex_code}}, which employs implicit Q-learning (IQL) \cite{142} as the backbone algorithm, serves as a reference for our experiments on the standard D4RL benchmark \cite{161}. All tasks tested are summarized in `Table 1' of `supplementary document.pdf'. Training is conducted over five different random seeds. The common hyper-parameters used for all baselines are detailed in `Table 2' of `supplementary document.pdf', aligning with those in the open-source code of PEX to ensure a fair comparison.

Several baselines are employed for comparison with our model: \textbf{(i) PEX}: offline-to-online RL using Policy Expansion, serving as the primary benchmark; \textbf{(ii) Offline}: uses only an offline policy with IQL, without online policy training during the online fine-tuning phase; \textbf{(iii) Buffer}: involves only online policy training using IQL with a replay buffer of $\mathrm{D}_{\text{off}} \cup \mathrm{D}_{\text{on}}$ without an offline policy \cite{160}. The contrasting characteristics of the Offline and Buffer models make them suitable for our evaluation. All baseline models are implemented using the open-source code of PEX, with IQL as the foundational algorithm. Online fine-tuning is performed across all models over 1M environmental steps.

To implement our model, we adhere to the format described in Section 3.1 of \cite{63}, labeled as XI-intra(100, informed, $p^{}$, $G$) (See `Appendix' of `supplementary document.pdf'), where `XI' denotes an intrinsic reward, `100' specifies an explore duration, `informed' identifies a trigger type, $p^{}$ represents the exploitation duration parameterized by a target rate $\rho$ (See `Table 3' in 'supplementary document.pdf'), and `$G$' indicates the greedy mode without a starting mode for exploration. In our implementation, we have made several modifications: the value of the target rate $\rho$ is adjusted to [0.0001, 0.9], differing from the values (0.1, 0.01, 0.001, 0.0001) used in \cite{63}. The intrinsic reward is replaced with an 'external reward,' specifically, an environmental reward used during online fine-tuning. The 'explore duration' is replaced with `$explore\_fixed\_steps$'. For triggering, `Homeo' is employed, focusing on the 'value promise' of $Q^{\text{off}}$.

\begin{figure*}
  \centering
  \subcaptionbox{\label{fig:antmaze-umaze-smoothing}}{
  \includegraphics[width=1.55in]{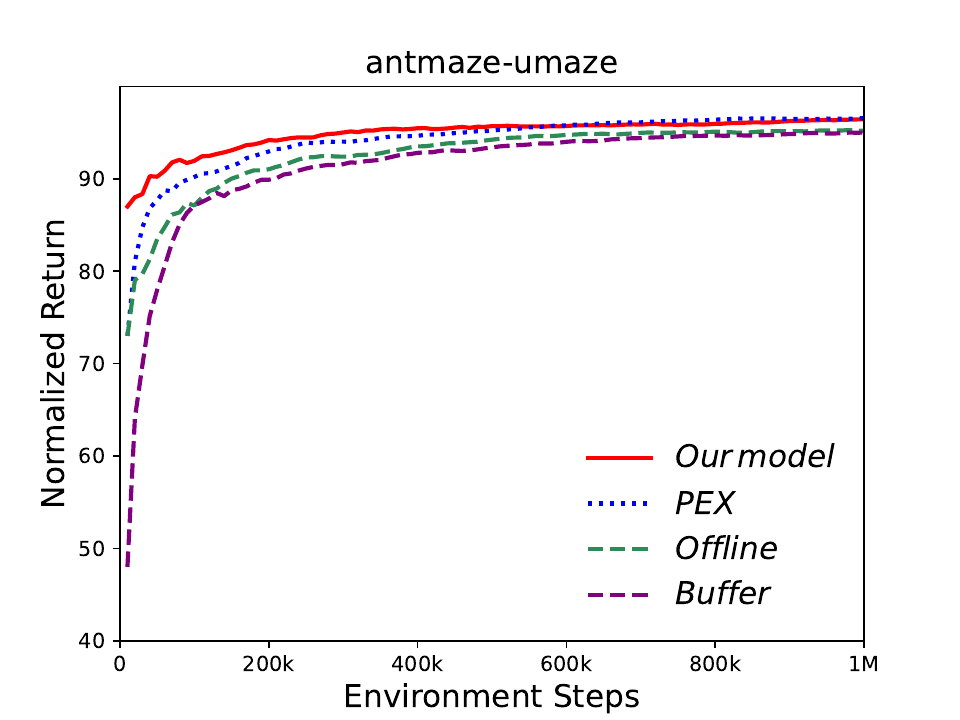}}
  \subcaptionbox{\label{fig:antmaze-umaze-diverse-smoothing}}{
  \includegraphics[width=1.55in]{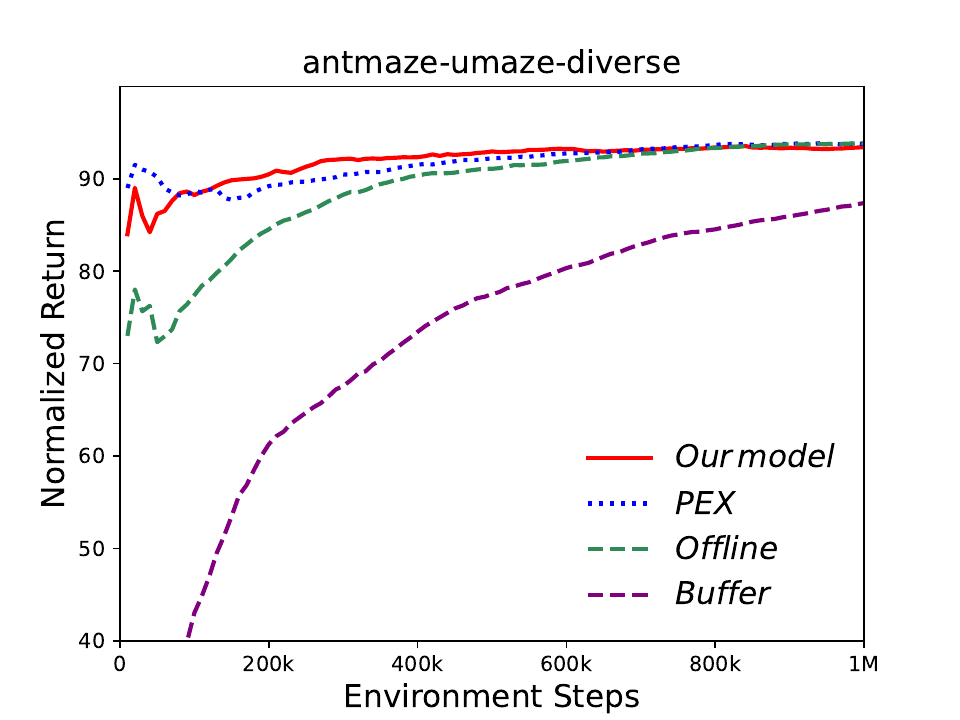}}%
  \subcaptionbox{\label{fig:antmaze-medium-play-smoothing}}{
  \includegraphics[width=1.55in]{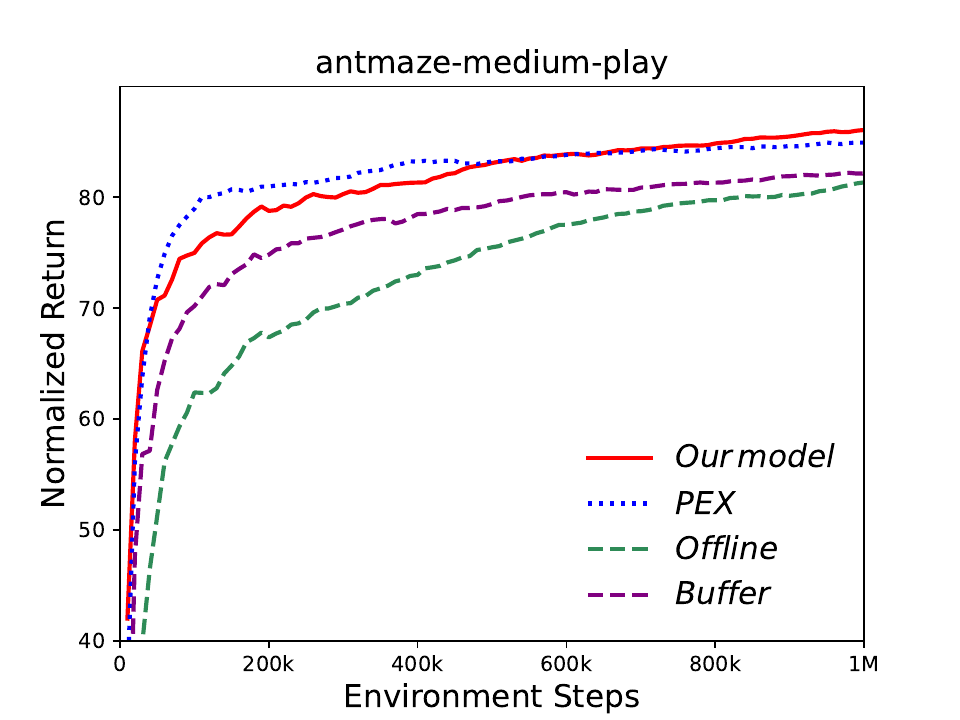}}

  \centering
  \bigskip
  \subcaptionbox{\label{fig:antmaze-medium-diverse-smoothing}}{
  \includegraphics[width=1.55in]{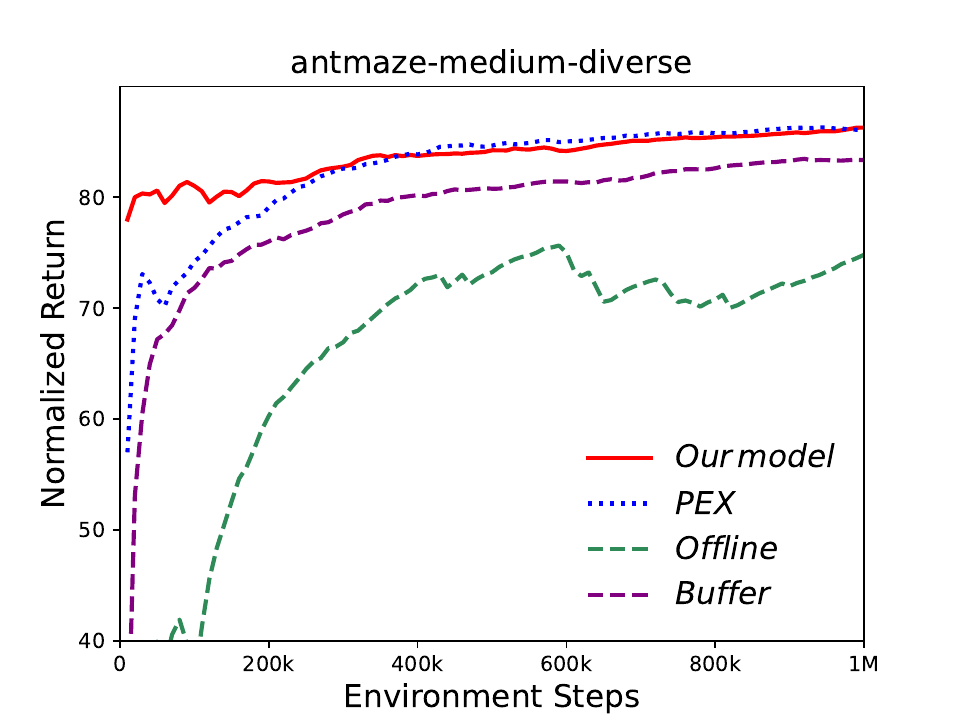}}%
  \subcaptionbox{\label{fig:antmaze-large-play-smoothing}}{  
  \includegraphics[width=1.55in]{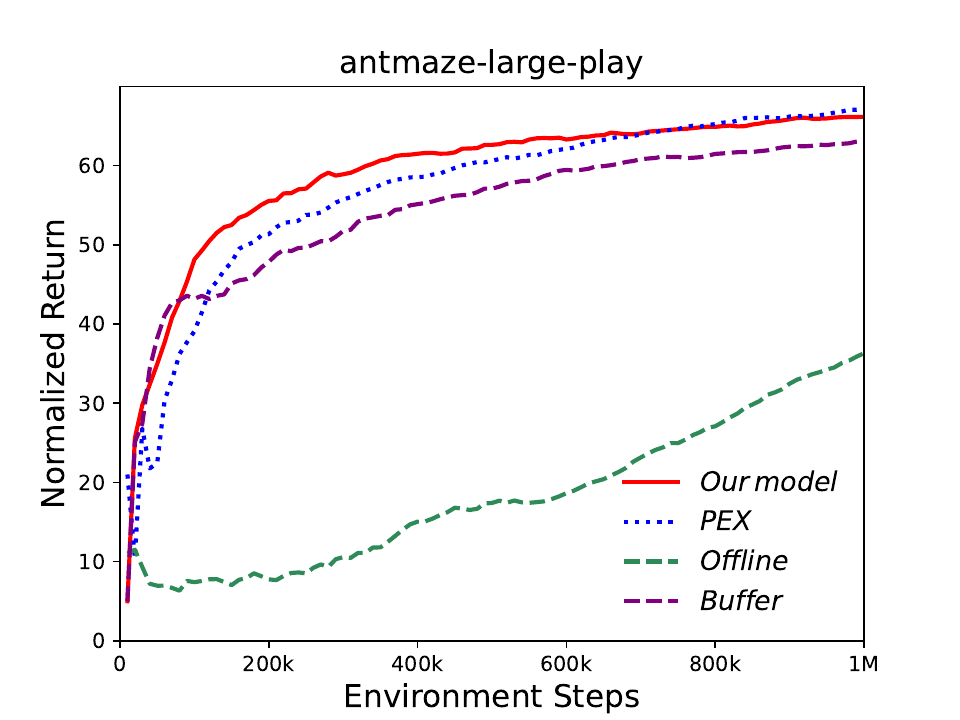}}%
  \subcaptionbox{\label{fig:antmaze-large-diverse-smoothing}}{
  \includegraphics[width=1.55in]{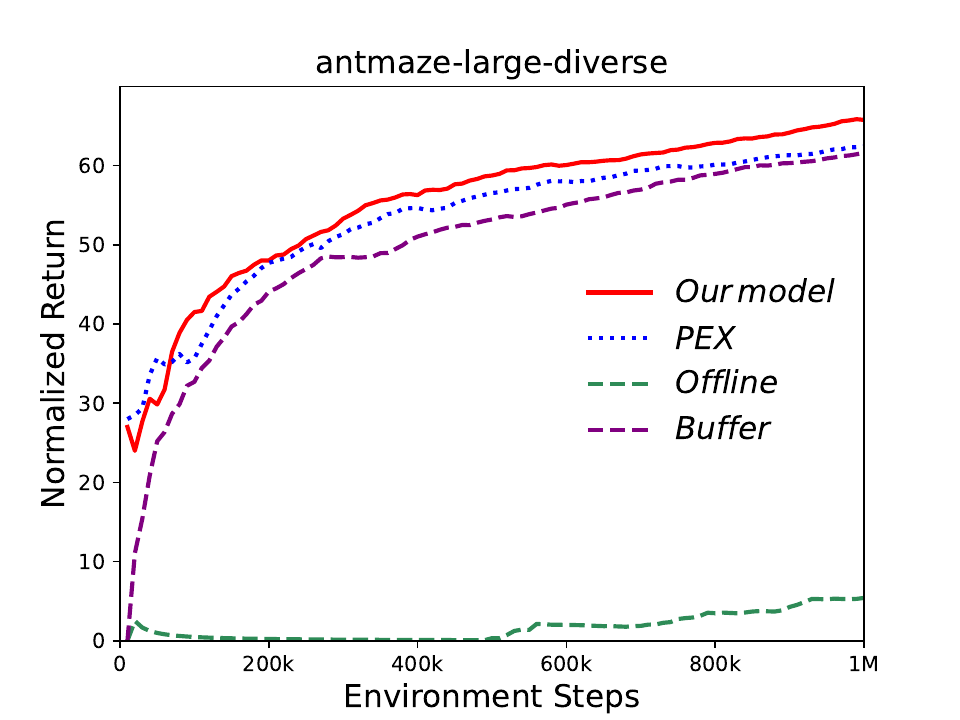}}

  \centering
  \bigskip
  \subcaptionbox{\label{fig:halfcheetah-random-smoothing}}{
  \includegraphics[width=1.55in]{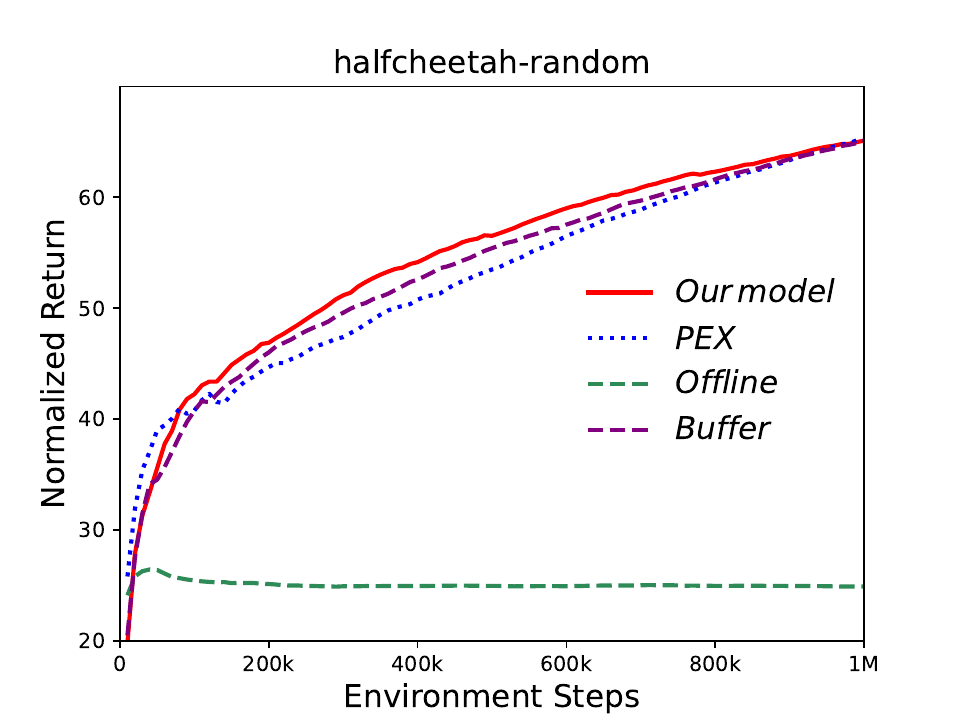}}%
  \subcaptionbox{\label{fig:hopper-random-smoothing}}{
  \includegraphics[width=1.55in]{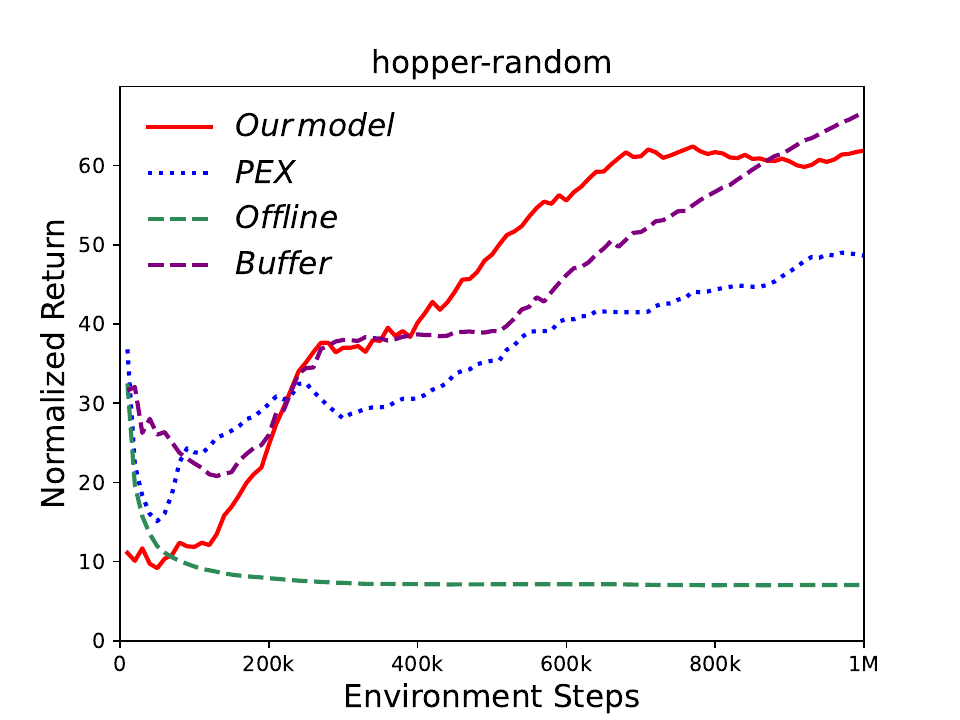}}
  \subcaptionbox{\label{fig:walker2d-random-smoothing}}{
  \includegraphics[width=1.55in]{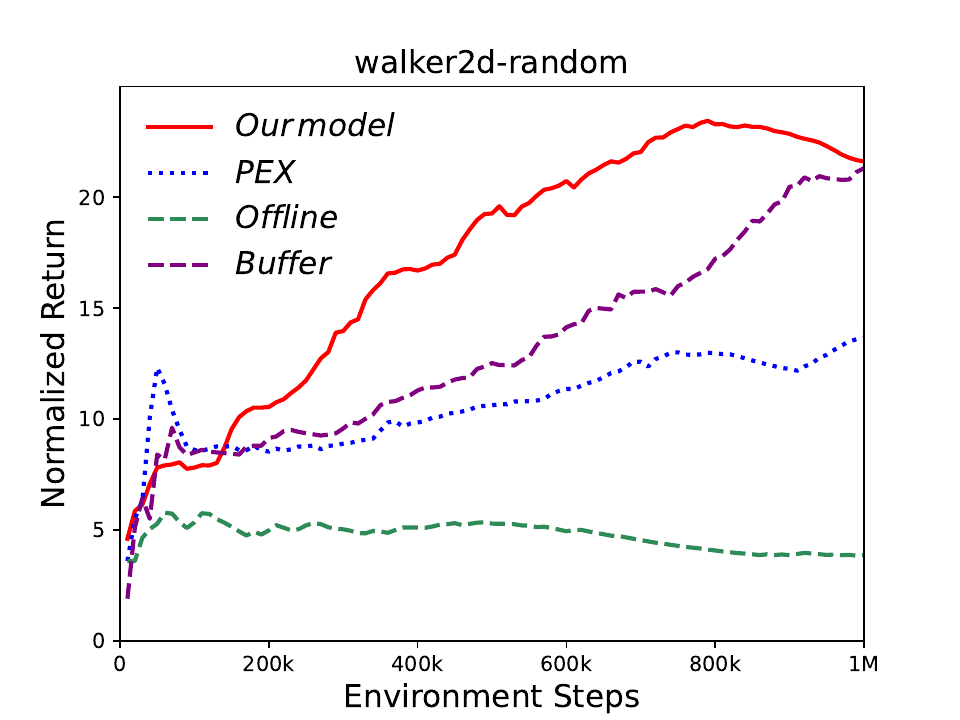}}%

  \centering
  \bigskip
  \subcaptionbox{\label{fig:halfcheetah-medium-smotthing}}{  
  \includegraphics[width=1.55in]{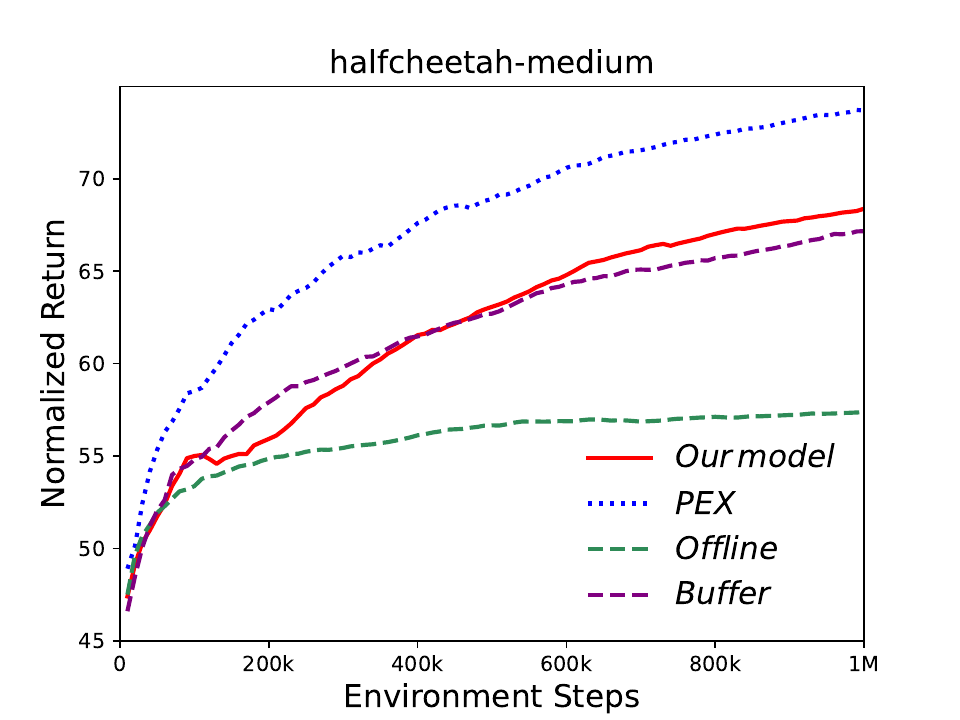}}%
  \subcaptionbox{\label{fig:hopper-medium-smoothing}}{
  \includegraphics[width=1.55in]{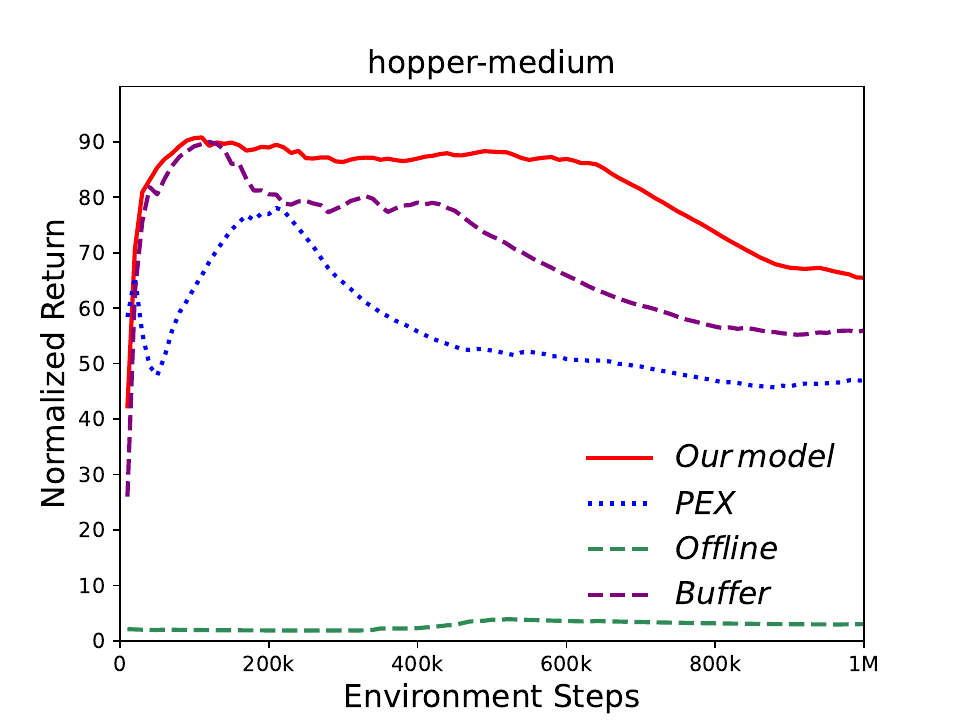}}
  \subcaptionbox{\label{fig:walker2d-medium_smoothing}}{
  \includegraphics[width=1.55in]{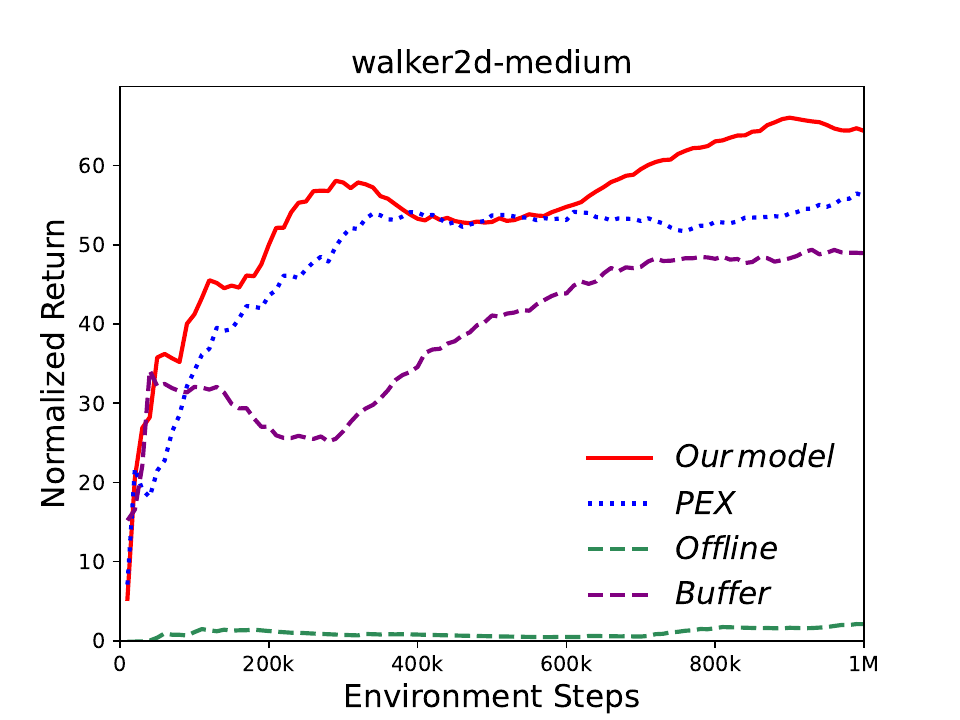}}%

  \centering
  \bigskip
  \subcaptionbox{\label{fig:halfcheetah-medium-replay-smoothing}}{
  \includegraphics[width=1.55in]{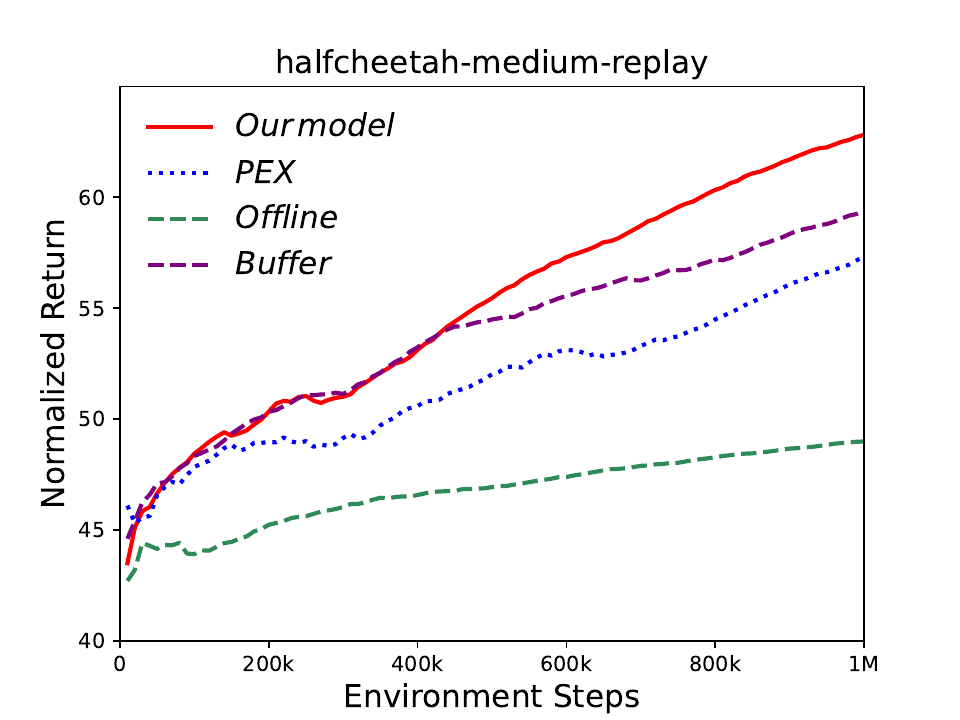}}
  \subcaptionbox{\label{fig:hopper-medium-replay_smoothing}}{
  \includegraphics[width=1.55in]{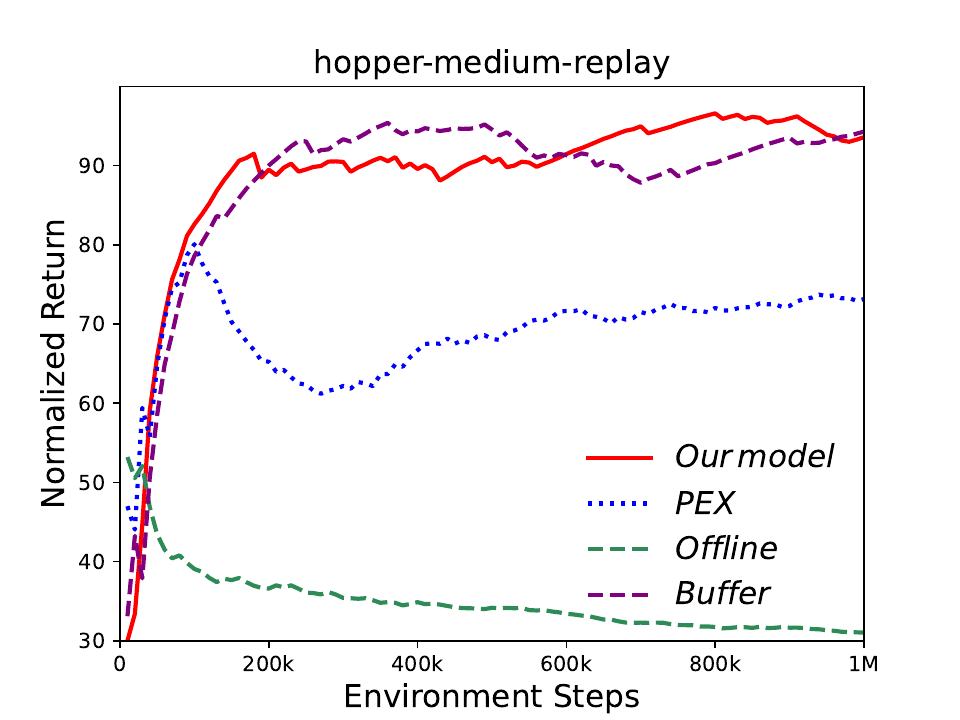}}%
  \subcaptionbox{\label{fig:walker2d-medium-replay-smoothing}}{  
  \includegraphics[width=1.55in]{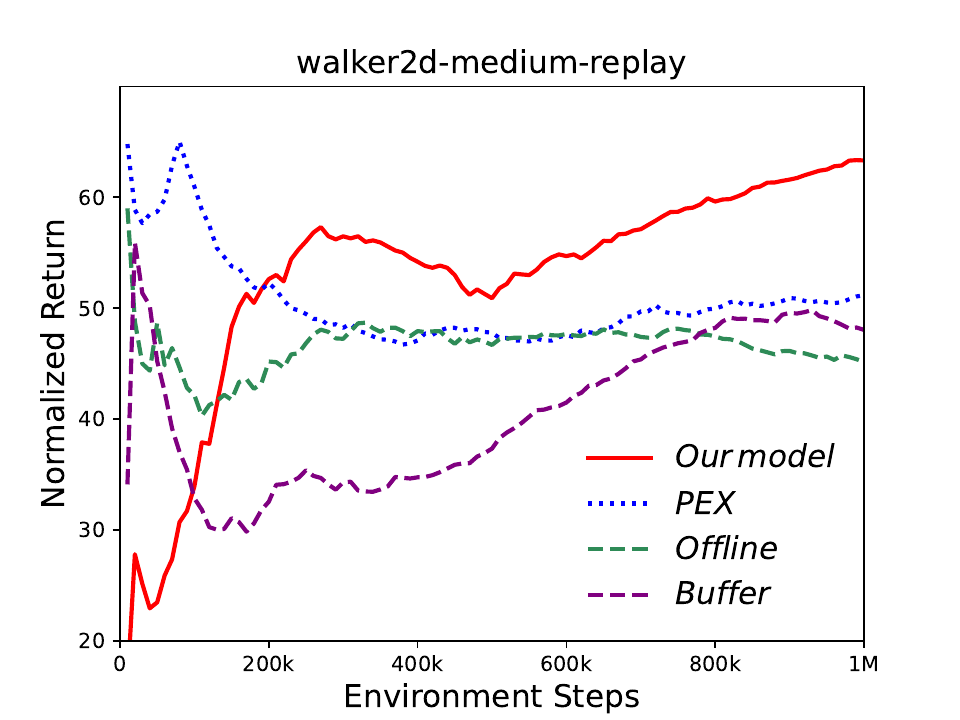}}%

  \caption{Normalized Return Curves of different methods, which are our model, PEX, Offline and Buffer, on benchmark tasks from D4RL. IQL is used for all methods as the backbone.}
  \label{fig:Result_15_smoothing}

\end{figure*}

\begin{figure*}
  \centering
  \subcaptionbox{\label{fig:antmaze-umaze_count}}{
  \includegraphics[width=1.55in]{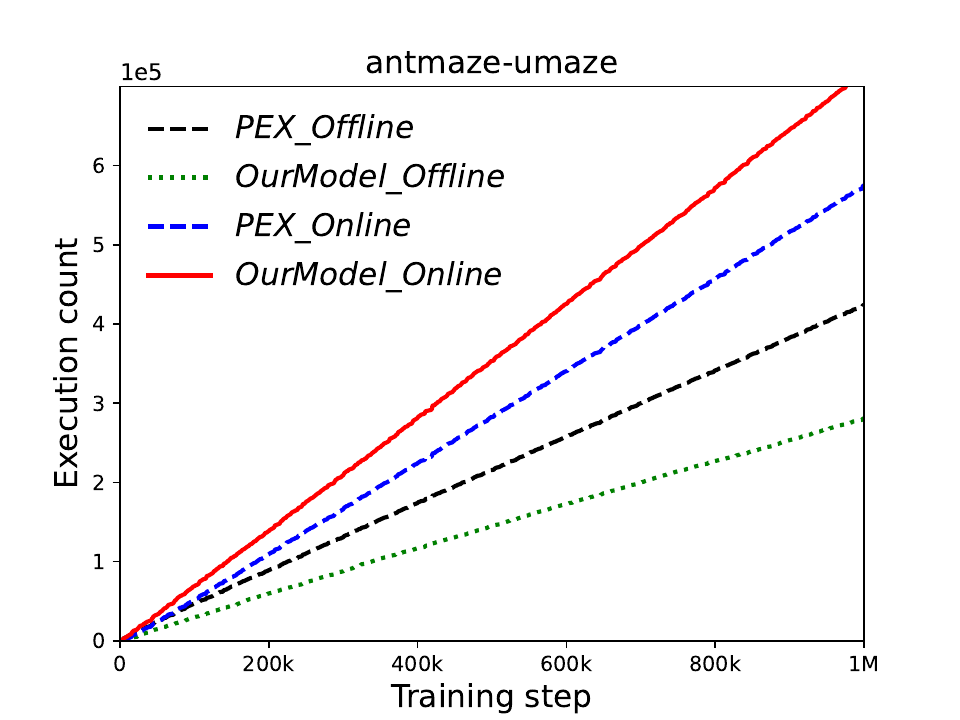}}
  \subcaptionbox{\label{fig:antmaze-umaze-diverse-count}}{
  \includegraphics[width=1.55in]{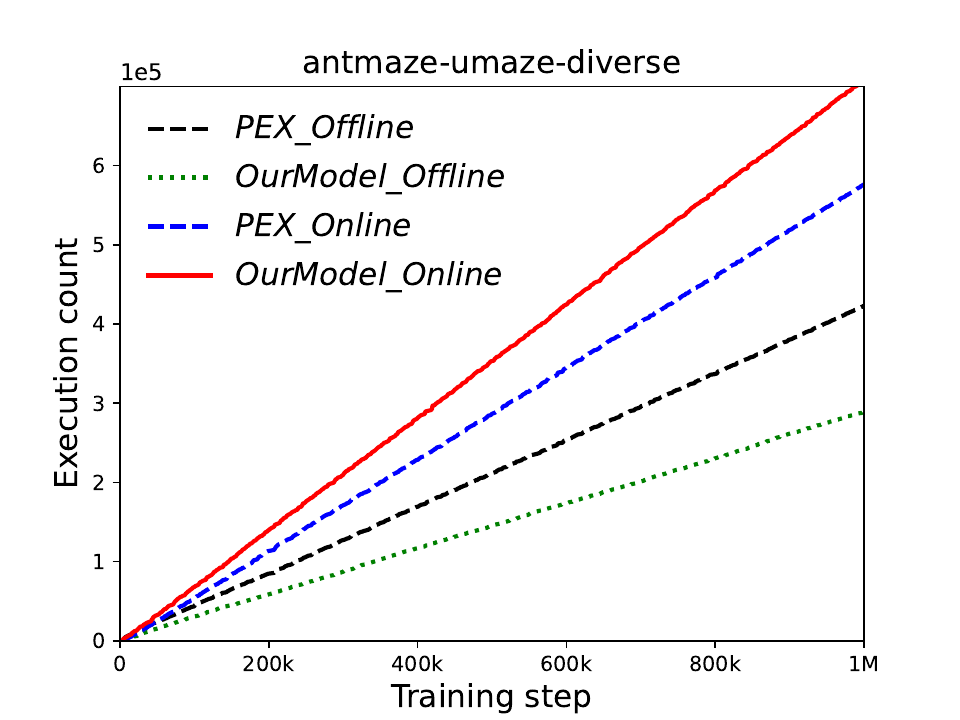}}%
  \subcaptionbox{\label{fig:antmaze-medium-play-count}}{
  \includegraphics[width=1.55in]{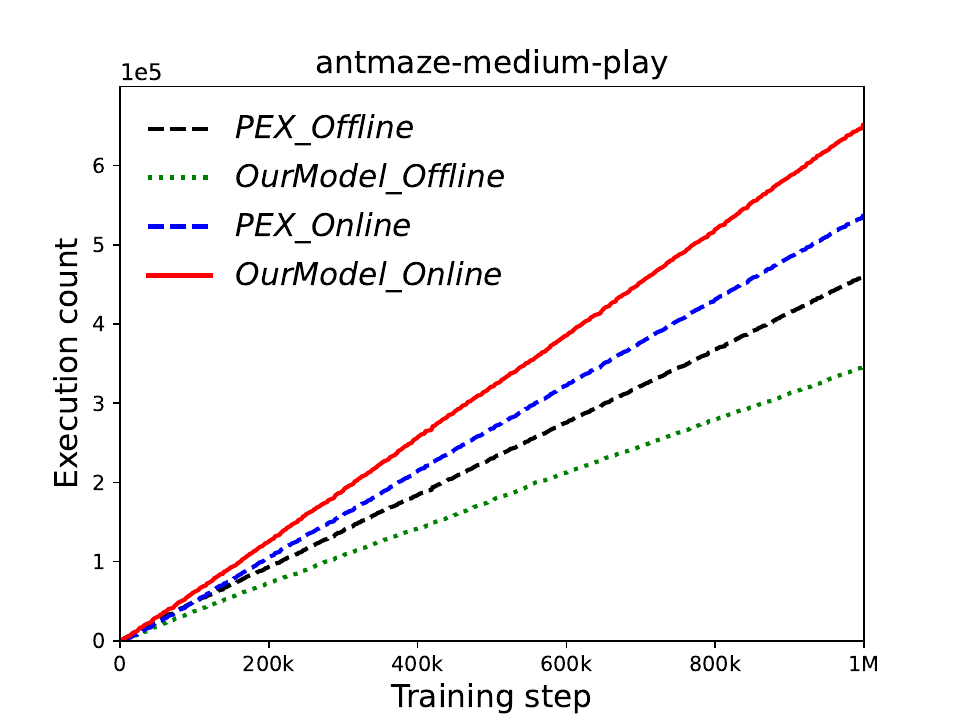}}
  
  \centering
  \bigskip
  \subcaptionbox{\label{fig:antmaze-medium-diverse-count}}{
  \includegraphics[width=1.55in]{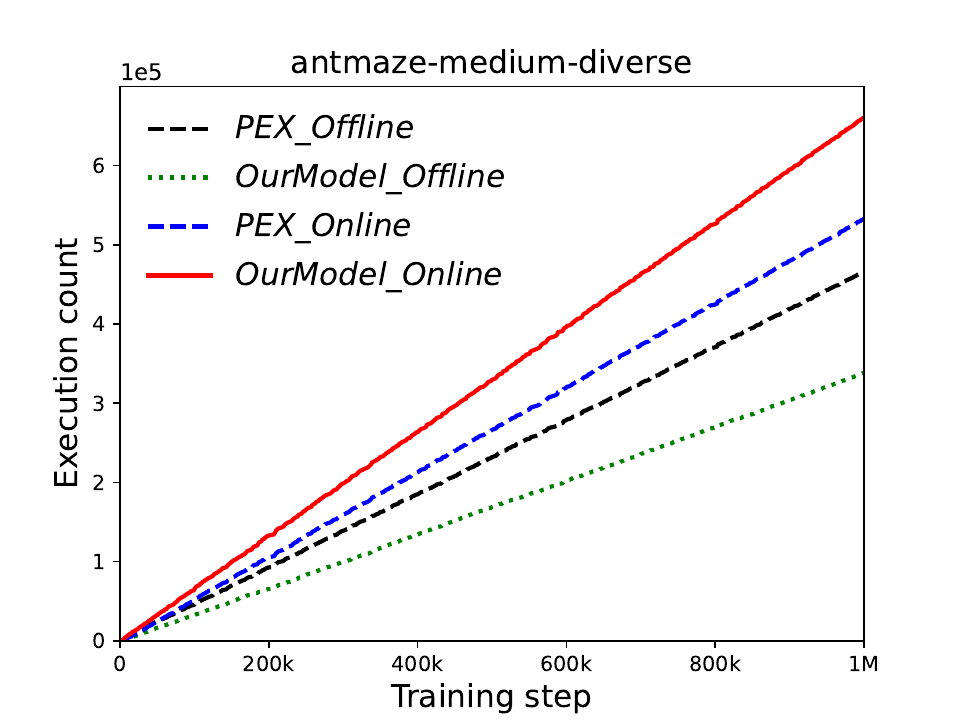}}%
  \subcaptionbox{\label{fig:antmaze-large-play-count}}{  
  \includegraphics[width=1.55in]{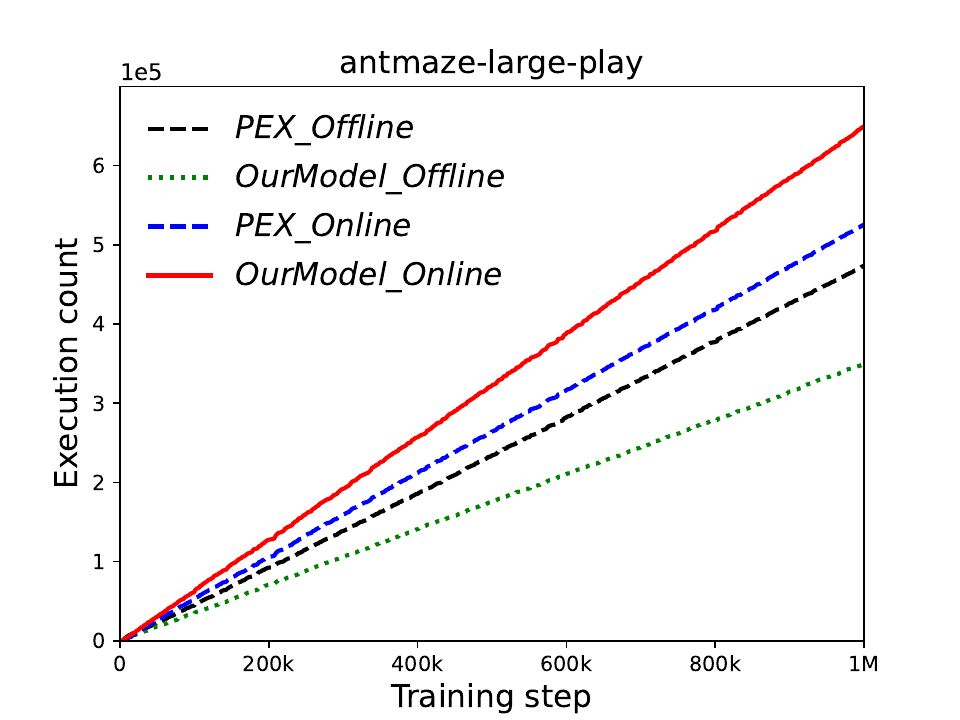}}%
  \subcaptionbox{\label{fig:antmaze-large-diverse-count}}{
  \includegraphics[width=1.55in]{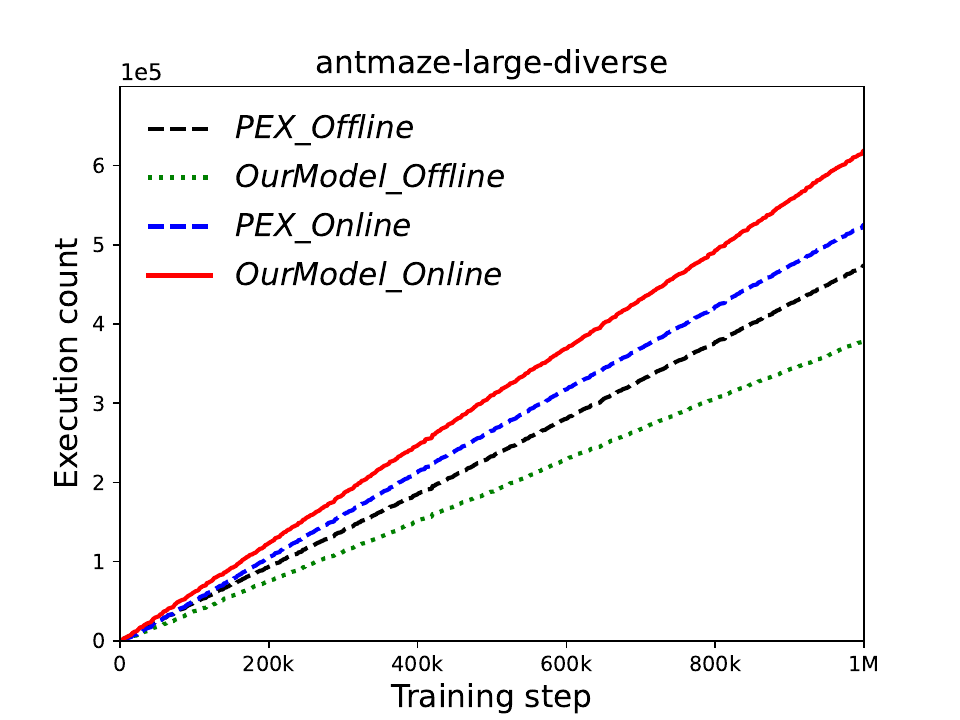}}

  \centering
  \bigskip
  \subcaptionbox{\label{fig:halfcheetah-random-count}}{
  \includegraphics[width=1.55in]{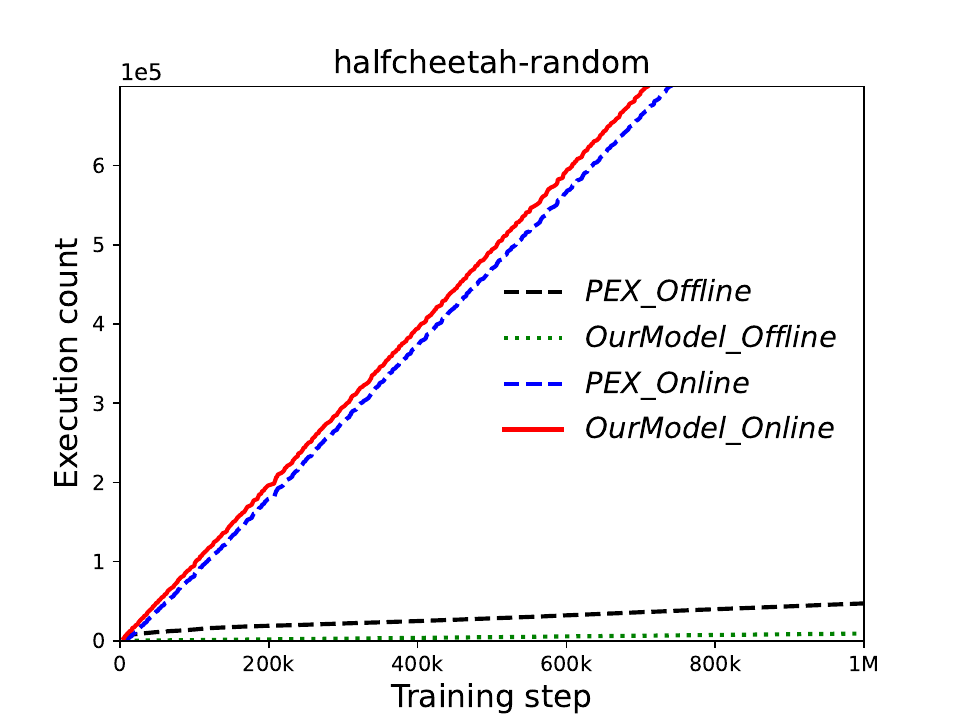}}%
  \subcaptionbox{\label{fig:hopper-random-count}}{
  \includegraphics[width=1.55in]{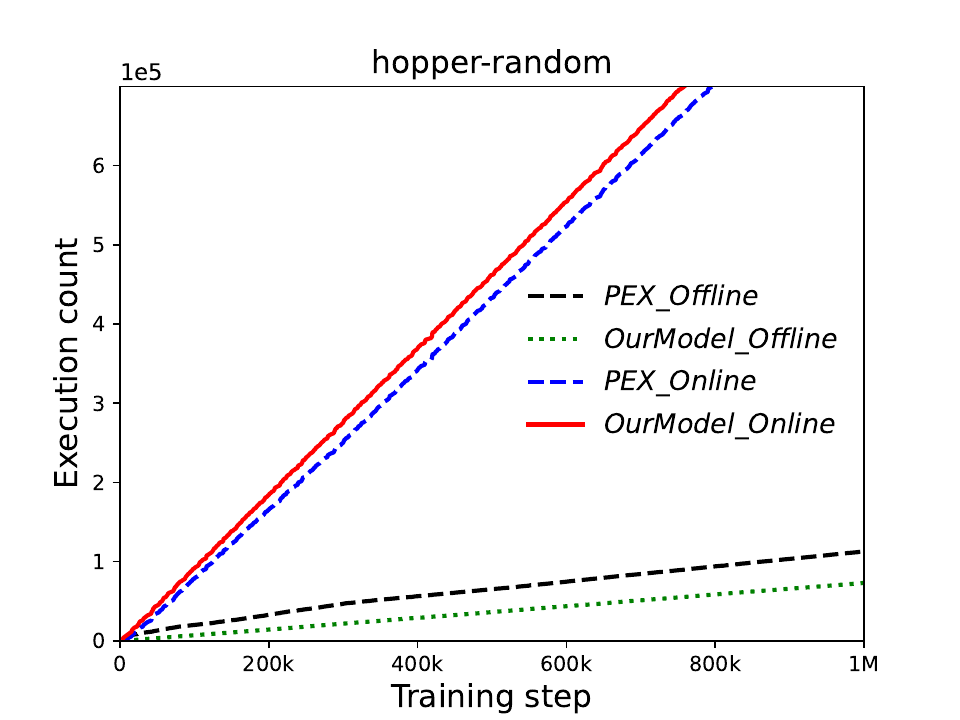}}
  \subcaptionbox{\label{fig:walker2d-random-count}}{
  \includegraphics[width=1.55in]{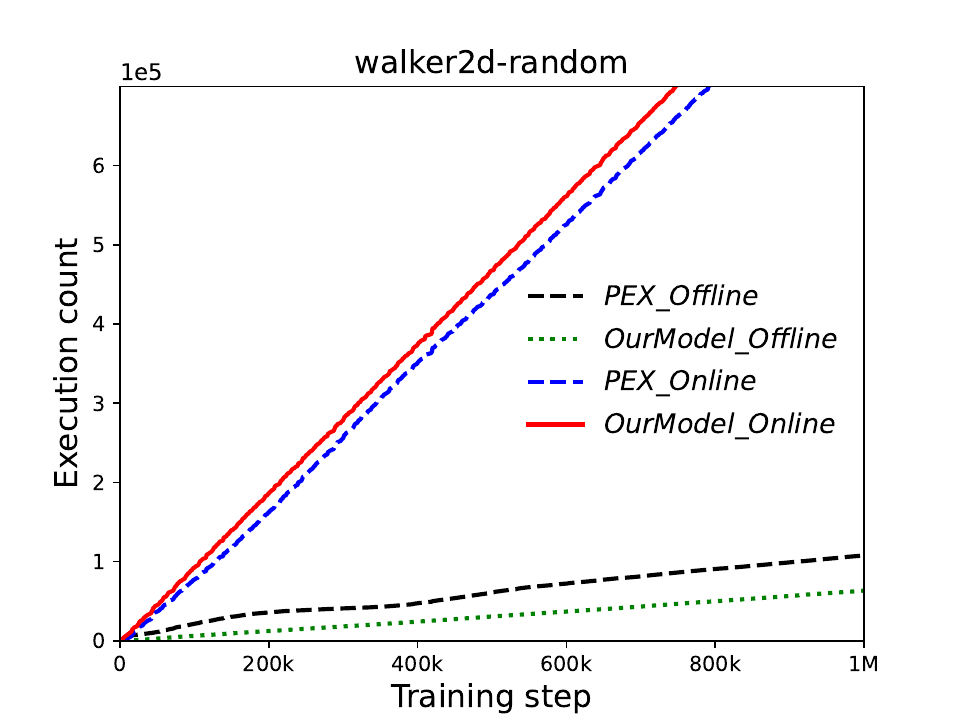}}%

  \centering
  \bigskip
  \subcaptionbox{\label{fig:halfcheetah-medium-count}}{  
  \includegraphics[width=1.55in]{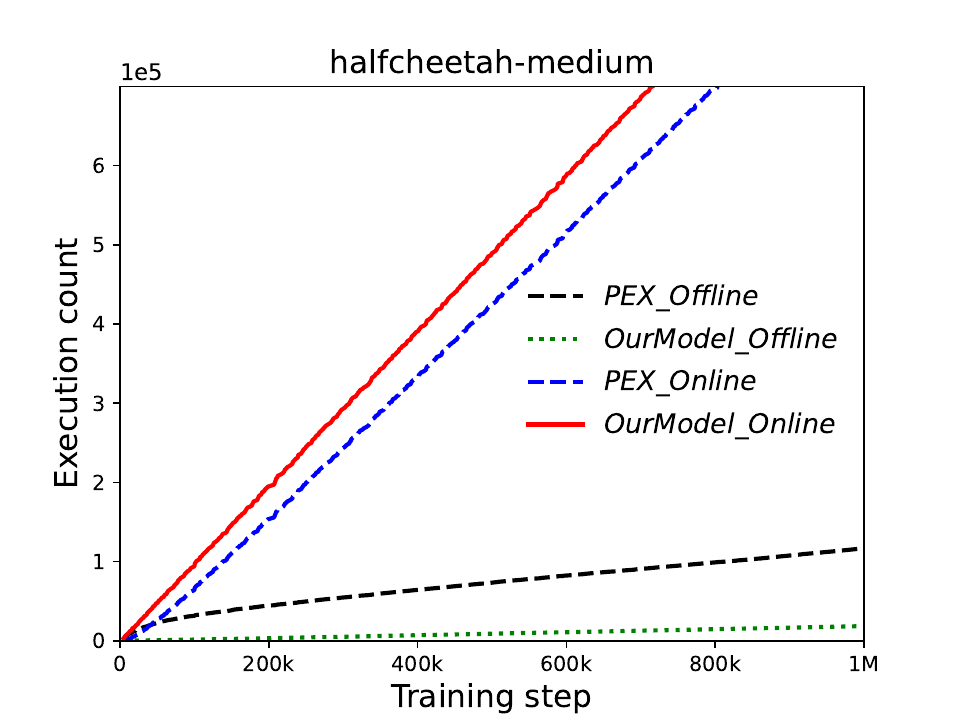}}%
  \subcaptionbox{\label{fig:hopper-medium-conut}}{
  \includegraphics[width=1.55in]{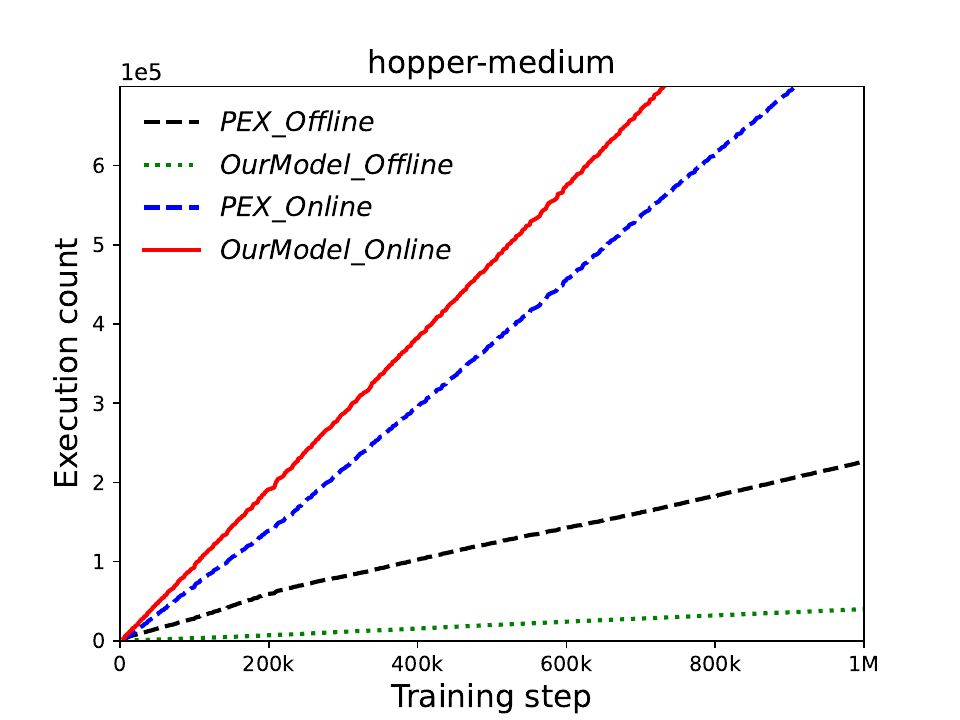}}
  \subcaptionbox{\label{fig:walker2d-medium_count}}{
  \includegraphics[width=1.55in]{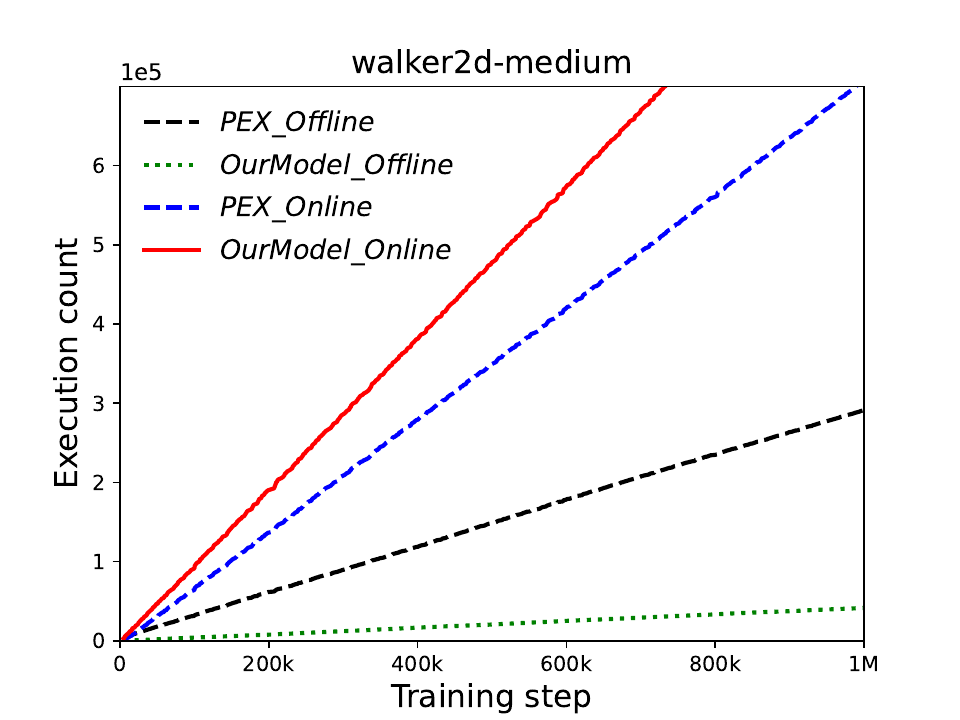}}%
  
  \centering
  \bigskip
  \subcaptionbox{\label{fig:halfcheetah-medium-replay-count}}{
  \includegraphics[width=1.55in]{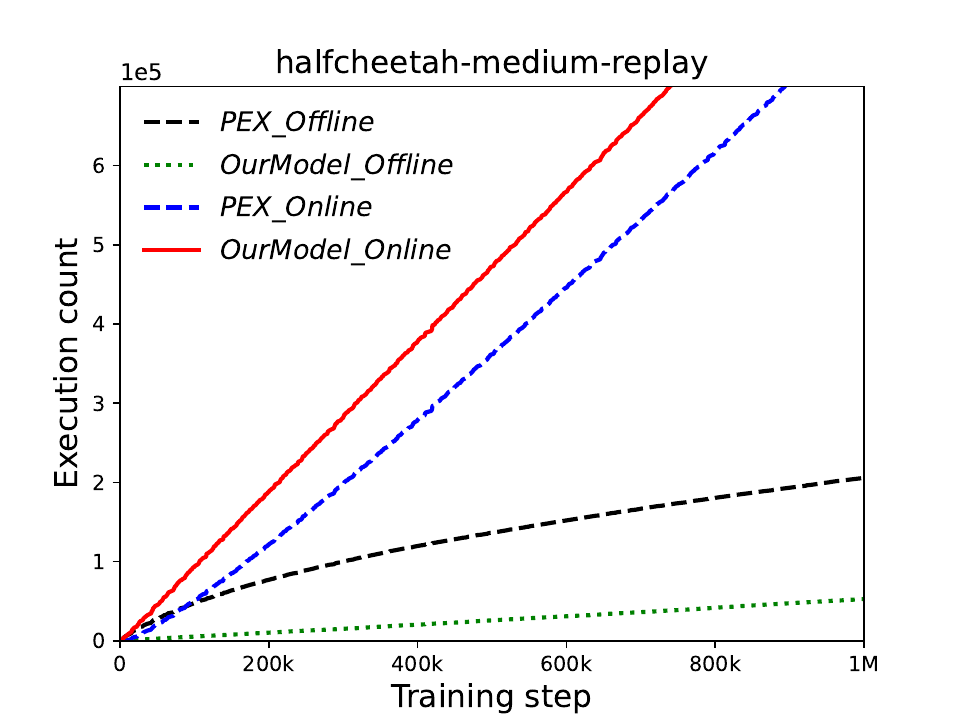}}
  \subcaptionbox{\label{fig:hopper-medium-replay_count}}{
  \includegraphics[width=1.55in]{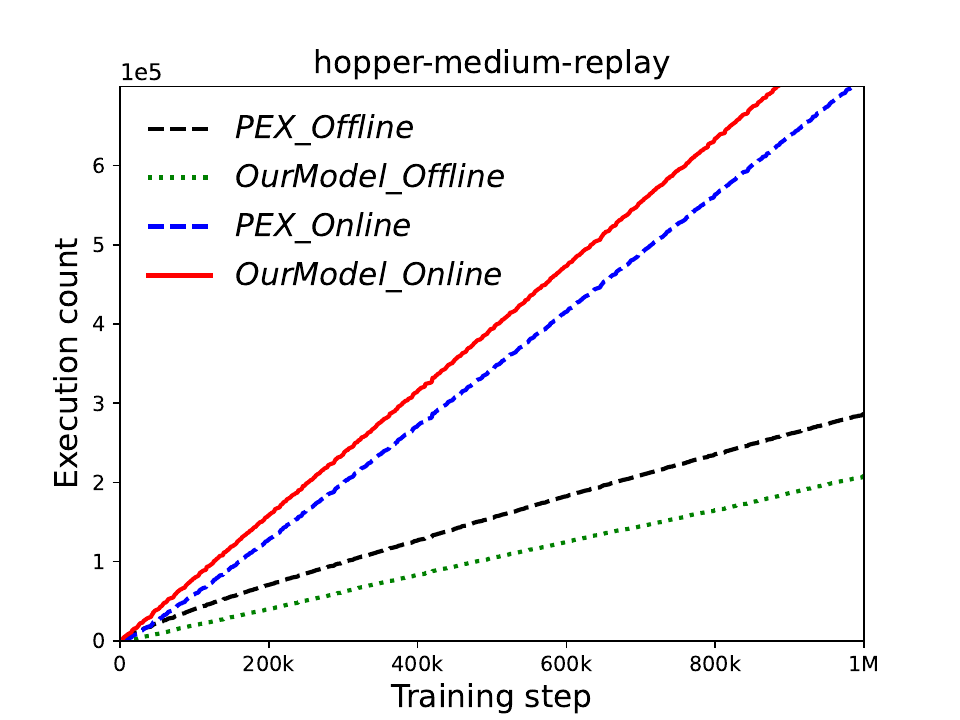}}%
  \subcaptionbox{\label{fig:walker2d-medium-replay-count}}{  
  \includegraphics[width=1.55in]{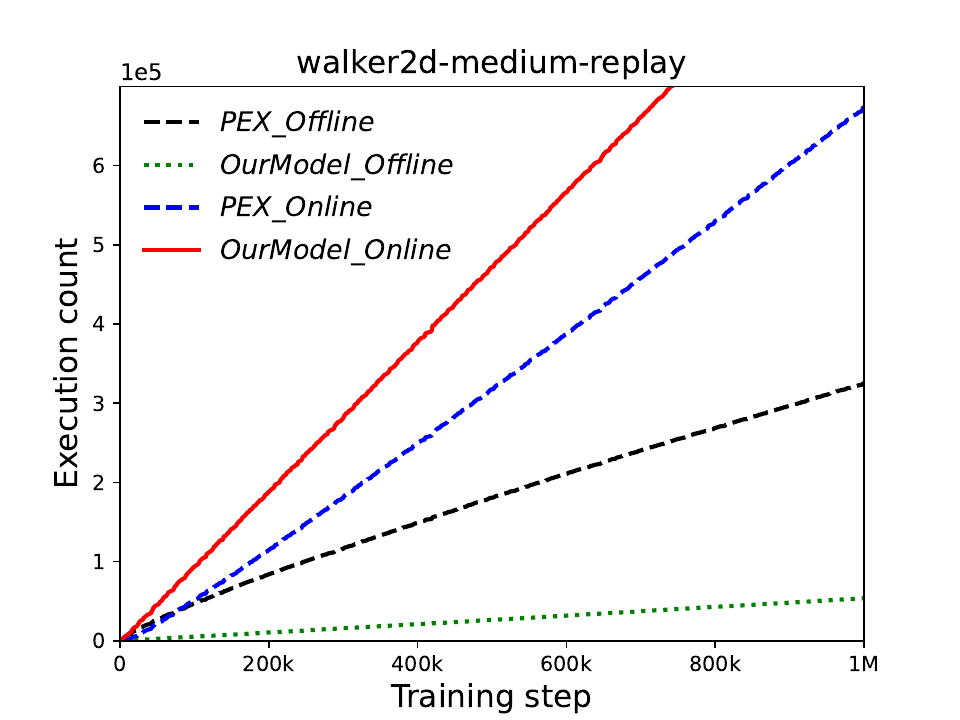}}%

  \caption{Execution count of our model and PEX on benchmark tasks from D4RL. The execution counts of offline policy and online policy of PEX or our model are referred to as $PEX\_Offline$ and $PEX\_Online$ or $OurModel\_Offline$ and $OurModel\_Online$, respectively.}
  \label{fig:Result_15_count}
\end{figure*}

The comparison process of our implementation unfolds as follows: First, an offline policy is pre-trained on a benchmark task from D4RL. Second, $\pi^{\text{off}}$  and $Q^{\text{off}}$, which are derived from this offline pre-training, are frozen and then utilized during online fine-tuning. The online policy is fine-tuned using three main parameters: $\rho$, `$explore\_fixed\_steps$' and `$update\_timestep$', tailored to a benchmark task from D4RL. To ensure a fair comparison, PEX, Offline, and Buffer are tested using the same pre-trained offline policy. The main algorithm of our non-monolithic exploration RL model during the fine-tuning phase is detailed in Algorithm \ref{alg:the_alg}.

There are two comparative aspects in our research. Firstly, our model is primarily compared with PEX. Secondly, all models experimented with in this study are compared to further research interests. In these experiments, our model demonstrates superior performance over PEX, Offline, and Buffer.

\subsection{The comparison between our model and PEX} \label{ourmodel_PEX}
Our research focuses on two key metrics: normalized return and execution count for both models. Initially, we analyze the normalized return across each task to compare the performance of the two models. This analysis will highlight the differences between our model, which employs non-monolithic exploration, and PEX, which utilizes policy expansion. Despite both models leveraging the same offline-to-online RL approach, our comparison aims to demonstrate their distinct performances without compromising the integrity of the offline RL policies.

Additionally, we analyze the number of times each policy—both offline and online—executes during the online fine-tuning phase for both models. By monitoring execution at every training step, this analysis helps illustrate the development of $\tilde{\pi}$ and estimate the contribution of each policy throughout the training process.

\subsubsection{Normalized return}
In the Antmaze environment, which features non-Markovian policies, sparse rewards, undirected, and multitask data, our model either matches or outperforms PEX in most cases, with the exception of `antmaze-medium-play'. Initially, our model faces challenges in `antmaze-umaze-diverse' and `antmaze-large-diverse' due to the immature online policy that undertakes exploration in datasets characterized by random goals and start locations. Paradoxically, the performance of our model surpasses that of PEX from an early stage, attributed to the sufficient exploration by the online policy. Notably, our model excels in `antmaze-large-diverse', which presents the most challenging conditions.

In environments such as HalfCheetah, Hopper, and Walker, which involve suboptimal agents and narrow data distributions, our model significantly outperforms PEX, with the exception of `halfcheetah-medium'. Initially, our model faces challenges due to exploration across most datasets. However, performance improves markedly from the early stages and is substantially better than that of PEX. Notably, our model demonstrates greater robustness in the latter stages, unlike PEX, whose performance tends to decline in `hopper-medium', `hopper-medium-replay', and `walker2d-medium-replay'.

\subsubsection{Execution count}
`Execution count' refers to the number of times each policy is executed at every training step. Our implementation code meticulously tracks the execution count of each policy at every training step. In the Antmaze environment, the discrepancy in execution count between the offline and online policies in PEX is smaller compared to our model. Our model significantly increases this difference, which correlates with enhanced performance. This indicates that PEX does not adequately reflect the current performance of $\tilde{\pi}$ during online fine-tuning, as evidenced by its overall performance metrics.

In the HalfCheetah, Hopper, and Walker environments, although there is a significant difference in the execution count between the offline and online policies in PEX, our model further increases this disparity. This observation suggests that the offline dataset does not substantially contribute to the performance of $\tilde{\pi}$ in PEX. In contrast, our model efficiently utilizes the offline policy to enhance the performance of $\tilde{\pi}$, even though the offline policy is executed less frequently. Meanwhile, the online policy is given more opportunities to execute within these domains, allowing it to further refine and solidify its own policy.

Across both models, the execution count of the online policy exceeds that of the offline policy, underscoring the reliance of $\tilde{\pi}$'s performance on the online policy. However, despite the lower execution count, the offline policy's contribution to $\tilde{\pi}$'s performance remains critical. The disparity between the execution counts of the offline and online policies varies with the dataset type, with a particularly pronounced difference observed in the 'random' dataset. Our model exhibits a larger discrepancy than PEX, even where the difference in PEX is substantial. This suggests that more extensive training of the online policy enhances performance on the dataset, while less frequent execution of the offline policy may be sufficient. Furthermore, in our model, the execution count for the offline policy is consistently lower than in PEX, and conversely, the count for the online policy is consistently higher (for further details, see `C Ablation Study' in `supplementary document.pdf'). 

\subsection{The comparison between our model and others} \label{ourmodel_others}
In Antmaze, while the performance of both Offline and Buffer trails behind our model and PEX, an interesting observation is that Buffer outperforms Offline in tasks such as `antmaze-medium-play', `antmaze-medium-diverse', `antmaze-large-play', and `antmaze-large-diverse'. This indicates that training the online policy is more crucial than relying solely on the pre-trained offline policy for these tasks. Notably, in `antmaze-large-diverse', the performance of the Offline model is particularly low, suggesting that the offline policy alone cannot significantly enhance the performance of $\tilde{\pi}$.

In HalfCheetah, Hopper, and Walker environments, Buffer outperforms PEX in all cases except for `halfcheetah-medium', `walker2d-medium', and `walker2d-medium-replay'. This underlines the significance of online fine-tuning, as demonstrated by Buffer’s superior performance compared to PEX. Our model, which shows even better performance than Buffer, effectively harnesses the synergy between the offline policy and the online policy, optimizing their combined impact on performance.

In three Antmaze tasks—`antmaze-umaze', `antmaze-umaze-diverse', and `antmaze-medium-diverse'—our model's performance is comparable to PEX. However, in `antmaze-medium-play' and `halfcheetah-medium', our model underperforms relative to PEX. Despite these outcomes, our model outperforms PEX in 10 out of all 15 tasks. Against Buffer, our model's performance is similar in `halfcheetah-random', `halfcheetah-medium', and `hopper-medium-replay'. Yet, it excels in 12 out of the 15 tasks evaluated, demonstrating a higher overall performance than Buffer.

\section{Discussion} \label{Discussion}

\subsection{Will $\pi^{\text{on}}$ be also suboptimal and even unsafe?}

Thanks to the adaptive role of $\pi^{\text{on}}$ throughout the stages of online fine-training (refer to Eq. \ref{entropy_equation_1} and Eq. \ref{entropy_equation_2}), $\pi^{\text{on}}$ is reliable. Except for `hopper-random' and `halfcheetah-medium', our model maintains competitive performance even in the late stages of training, demonstrating its effectiveness across various experiments.

\subsection{\textcolor{black}{The effect of non-monolithic exploration methodology for offline-to-online RL}}
As depicted in Fig. \ref{fig:Result_15_smoothing}, PEX faces challenges in ensuring sufficient training for the online policy. The execution counts of the offline and online policies in PEX are higher and lower, respectively, compared to those in our model. This suggests that PEX's design does not effectively manage the training of the online policy during the online fine-tuning phase. In contrast, our model provides adequate training for the online policy, effectively compensating for the relatively limited knowledge from the offline policy, which was pre-trained during the offline phase. This more robust approach in handling online policy training is further supported by the data in table \ref{Comparison_ourmodel_pex}.

\subsection{The efficient usage of offline policy during online fine-tuning}
Fig. \ref{fig:Result_15_count} demonstrates that our model utilizes the offline policy more effectively than PEX, despite having a smaller execution count for the offline policy. This efficient use of the offline policy significantly enhances the performance of $\tilde{\pi}$, optimizing the balance between policy use and overall effectiveness.

\subsection{How can the method be translated to a wider real-world problems?}
The application of our methodology to the distributed model of federated learning can be envisioned. In this setup, some local models might operate using an unmodified off-policy, while others adopt an on-policy approach, varying key parameters accordingly. The global model could then leverage a suitable mode-switching controller to select the optimal policy from among all the local models, effectively integrating diverse learning strategies to enhance overall performance.

\subsection{The considerations of further research}
Currently, the online fine-tuning process requires manual and fixed adjustment of three key parameters: $\rho$, `$explore\_fixed\_steps$', and `$update\_timestep$'for the online policy. However, we hypothesize that the performance of offline-to-online RL could be significantly improved if these parameters were dynamically adjusted by the agent itself. In this proposed scenario, a parameter controller such as 'Meta RL' would modulate these key parameters. A crucial aspect for effective offline-to-online RL, especially when the offline policy remains unmodified, involves how the execution times for each policy—offline or online—are adjusted.

This research primarily focuses on offline-to-online reinforcement learning (RL) using an unmodified offline policy. Consequently, PEX, which is the current state-of-the-art offline-to-online RL algorithm that leverages an unmodified offline policy, is chosen as the main reference for comparison. Additionally, we plan to extend our research to include comparisons with offline-to-online RL models based on the unified framework, where the online policy operates based on the parameters of the offline policy.

\section{Conclusion} \label{Conclusion}

In summary, this research introduces a novel non-monolithic exploration methodology for offline-to-online reinforcement learning (RL), which effectively utilizes an unmodified offline policy. This approach significantly enhances the agent's ability to leverage flexibility and generalization in downstream tasks, thereby improving performance across various scenarios.

\bibliographystyle{splncs04}
\bibliography{6813}

\begin{thebibliography}{10}
\providecommand{\url}[1]{\texttt{#1}}
\providecommand{\urlprefix}{URL }
\providecommand{\doi}[1]{https://doi.org/#1}

\bibitem{155}
An, G., Moon, S., Kim, J.H., Song, H.O.: Uncertainty-based offline reinforcement learning with diversified q-ensemble. Advances in neural information processing systems  \textbf{34},  7436--7447 (2021)

\bibitem{136}
Ashvin, N., Murtaza, D., Abhishek, G., Sergey, L.: Accelerating online reinforcement learning with offline datasets. CoRR, vol. abs/2006.09359  (2020)

\bibitem{128}
Ebitz, R.B., Sleezer, B.J., Jedema, H.P., Bradberry, C.W., Hayden, B.Y.: Tonic exploration governs both flexibility and lapses. PLoS computational biology  \textbf{15}(11),  e1007475 (2019)

\bibitem{161}
Fu, J., Kumar, A., Nachum, O., Tucker, G., Levine, S.: D4rl: Datasets for deep data-driven reinforcement learning. arXiv preprint arXiv:2004.07219  (2020)

\bibitem{147}
Fujimoto, S., Meger, D., Precup, D.: Off-policy deep reinforcement learning without exploration. In: International conference on machine learning. pp. 2052--2062. PMLR (2019)

\bibitem{127}
Gershman, S.J., Tzovaras, B.G.: Dopaminergic genes are associated with both directed and random exploration. Neuropsychologia  \textbf{120},  97--104 (2018)

\bibitem{137}
Guo, S., Sun, Y., Hu, J., Huang, S., Chen, H., Piao, H., Sun, L., Chang, Y.: A simple unified uncertainty-guided framework for offline-to-online reinforcement learning. arXiv preprint arXiv:2306.07541  (2023)

\bibitem{151}
Kostrikov, I., Fergus, R., Tompson, J., Nachum, O.: Offline reinforcement learning with fisher divergence critic regularization. In: International Conference on Machine Learning. pp. 5774--5783. PMLR (2021)

\bibitem{142}
Kostrikov, I., Nair, A., Levine, S.: Offline reinforcement learning with implicit q-learning. arXiv preprint arXiv:2110.06169  (2021)

\bibitem{148}
Kumar, A., Fu, J., Soh, M., Tucker, G., Levine, S.: Stabilizing off-policy q-learning via bootstrapping error reduction. Advances in Neural Information Processing Systems  \textbf{32} (2019)

\bibitem{141}
Kumar, A., Zhou, A., Tucker, G., Levine, S.: Conservative q-learning for offline reinforcement learning. Advances in Neural Information Processing Systems  \textbf{33},  1179--1191 (2020)

\bibitem{84}
Laskin, M., Yarats, D., Liu, H., Lee, K., Zhan, A., Lu, K., Cang, C., Pinto, L., Abbeel, P.: Urlb: Unsupervised reinforcement learning benchmark. arXiv preprint arXiv:2110.15191  (2021)

\bibitem{130}
Lee, S., Seo, Y., Lee, K., Abbeel, P., Shin, J.: Offline-to-online reinforcement learning via balanced replay and pessimistic q-ensemble. In: Conference on Robot Learning. pp. 1702--1712. PMLR (2022)

\bibitem{129}
Levine, S., Kumar, A., Tucker, G., Fu, J.: Offline reinforcement learning: Tutorial, review, and perspectives on open problems. arXiv preprint arXiv:2005.01643  (2020)

\bibitem{146}
Levine, S., Kumar, A., Tucker, G., Fu, J.: Offline reinforcement learning: Tutorial, review, and perspectives on open problems. arXiv preprint arXiv:2005.01643  (2020)

\bibitem{131}
Mao, Y., Wang, C., Wang, B., Zhang, C.: Moore: Model-based offline-to-online reinforcement learning. arXiv preprint arXiv:2201.10070  (2022)

\bibitem{157}
Nair, A., Gupta, A., Dalal, M., Levine, S.: Awac: Accelerating online reinforcement learning with offline datasets. arXiv preprint arXiv:2006.09359  (2020)

\bibitem{63}
Pislar, M., Szepesvari, D., Ostrovski, G., Borsa, D., Schaul, T.: When should agents explore? arXiv preprint arXiv:2108.11811  (2021)

\bibitem{145}
Prudencio, R.F., Maximo, M.R., Colombini, E.L.: A survey on offline reinforcement learning: Taxonomy, review, and open problems. IEEE Transactions on Neural Networks and Learning Systems  (2023)

\bibitem{162}
Schwarzer, M., Rajkumar, N., Noukhovitch, M., Anand, A., Charlin, L., Hjelm, R.D., Bachman, P., Courville, A.C.: Pretraining representations for data-efficient reinforcement learning. Advances in Neural Information Processing Systems  \textbf{34},  12686--12699 (2021)

\bibitem{134}
Siegel, N.Y., Springenberg, J.T., Berkenkamp, F., Abdolmaleki, A., Neunert, M., Lampe, T., Hafner, R., Heess, N., Riedmiller, M.: Keep doing what worked: Behavioral modelling priors for offline reinforcement learning. arXiv preprint arXiv:2002.08396  (2020)

\bibitem{135}
Singh, A., Yu, A., Yang, J., Zhang, J., Kumar, A., Levine, S.: Cog: Connecting new skills to past experience with offline reinforcement learning. arXiv preprint arXiv:2010.14500  (2020)

\bibitem{64}
Turrigiano, G.G., Nelson, S.B.: Homeostatic plasticity in the developing nervous system. Nature reviews neuroscience  \textbf{5}(2),  97--107 (2004)

\bibitem{160}
Vecerik, M., Hester, T., Scholz, J., Wang, F., Pietquin, O., Piot, B., Heess, N., Rothörl, T., Lampe, T., Riedmiller, M.: Leveraging demonstrations for deep reinforcement learning on robotics problems with sparse rewards. arXiv preprint arXiv:1707.08817  (2017)

\bibitem{154}
Wu, Y., Zhai, S., Srivastava, N., Susskind, J., Zhang, J., Salakhutdinov, R., Goh, H.: Uncertainty weighted actor-critic for offline reinforcement learning. arXiv preprint arXiv:2105.08140  (2021)

\bibitem{156}
Xie, Z., Lin, Z., Li, J., Li, S., Ye, D.: Pretraining in deep reinforcement learning: A survey. arXiv preprint arXiv:2211.03959  (2022)

\bibitem{158}
Yu, T., Kumar, A., Rafailov, R., Rajeswaran, A., Levine, S., Finn, C.: Combo: Conservative offline model-based policy optimization. Advances in neural information processing systems  \textbf{34},  28954--28967 (2021)

\bibitem{132}
Zhang, H., Xu, W., Yu, H.: Policy expansion for bridging offline-to-online reinforcement learning. arXiv preprint arXiv:2302.00935  (2023)

\bibitem{163}
Zhang, Q., Peng, Z., Zhou, B.: Learning to drive by watching youtube videos: Action-conditioned contrastive policy pretraining. In: European Conference on Computer Vision. pp. 111--128. Springer (2022)

\bibitem{139}
Zheng, H., Luo, X., Wei, P., Song, X., Li, D., Jiang, J.: Adaptive policy learning for offline-to-online reinforcement learning. arXiv preprint arXiv:2303.07693  (2023)

\bibitem{133}
Zheng, Q., Zhang, A., Grover, A.: Online decision transformer. In: international conference on machine learning. pp. 27042--27059. PMLR (2022)

\bibitem{159}
Zhu, D., Wang, Y., Schmidhuber, J., Elhoseiny, M.: Guiding online reinforcement learning with action-free offline pretraining. arXiv preprint arXiv:2301.12876  (2023)

\end{thebibliography}

\end{document}